\pdfoutput=1

\documentclass[11pt]{article}

\usepackage[]{acl}  
\usepackage{nert}
\usepackage{todonotes}

\usepackage{times}
\usepackage{latexsym}

\usepackage[utf8]{inputenc}

\newcommand{\Dataset}{\textbf{ConQRet}}
\newcommand{\Task}{\textbf{RAArg}}
\usepackage{array}
\usepackage{caption}
\usepackage{booktabs}
\usepackage{multirow}
\usepackage{array}
\usepackage[normalem]{ulem} 

\title{\Dataset: Benchmarking Fine-Grained Evaluation of Retrieval Augmented Argumentation with LLM Judges}

\author{Kaustubh D. Dhole, Kai Shu, Eugene Agichtein \\
Department of Computer Science \\
  Emory University, Atlanta USA \\
  \eml{\{kaustubh.dhole, kai.shu, eugene.agichtein\}@emory.edu}
}
\date{}

\begin{document}
\maketitle
\begin{abstract}

Computational argumentation, which involves generating answers or summaries for controversial topics like abortion bans and vaccination, has become increasingly important in today's polarized environment. Sophisticated LLM capabilities offer the potential to provide nuanced, evidence-based answers to such questions through Retrieval-Augmented Argumentation (\Task), leveraging real-world evidence for high-quality, grounded arguments. However, evaluating~\Task{} remains challenging, as human evaluation is costly and difficult for complex, lengthy answers on complicated topics. At the same time, re-using existing argumentation datasets is no longer sufficient, as they lack long, complex arguments and realistic evidence from potentially misleading sources, limiting holistic evaluation of retrieval effectiveness and argument quality. To address these gaps, we investigate automated evaluation methods using multiple fine-grained LLM judges, providing better and more interpretable assessments than traditional single-score metrics and even previously reported human crowdsourcing. To validate the proposed techniques, we introduce~\Dataset{}, a new benchmark featuring long and complex~\textbf{human-authored arguments} on debated topics, grounded in~\textbf{real-world websites}, allowing an exhaustive evaluation across retrieval effectiveness, argument quality, and groundedness. We validate our LLM Judges on a prior dataset and the new~\Dataset{} benchmark. Our proposed LLM Judges and the~\Dataset{} benchmark can enable rapid progress in computational argumentation and can be naturally extended to other complex retrieval-augmented generation tasks.
\end{abstract}

\section{Introduction}
\label{sec:intro}

\begin{figure*}[h]
    \centering
    \begin{subfigure}{\textwidth}
        \includegraphics[width=\linewidth]{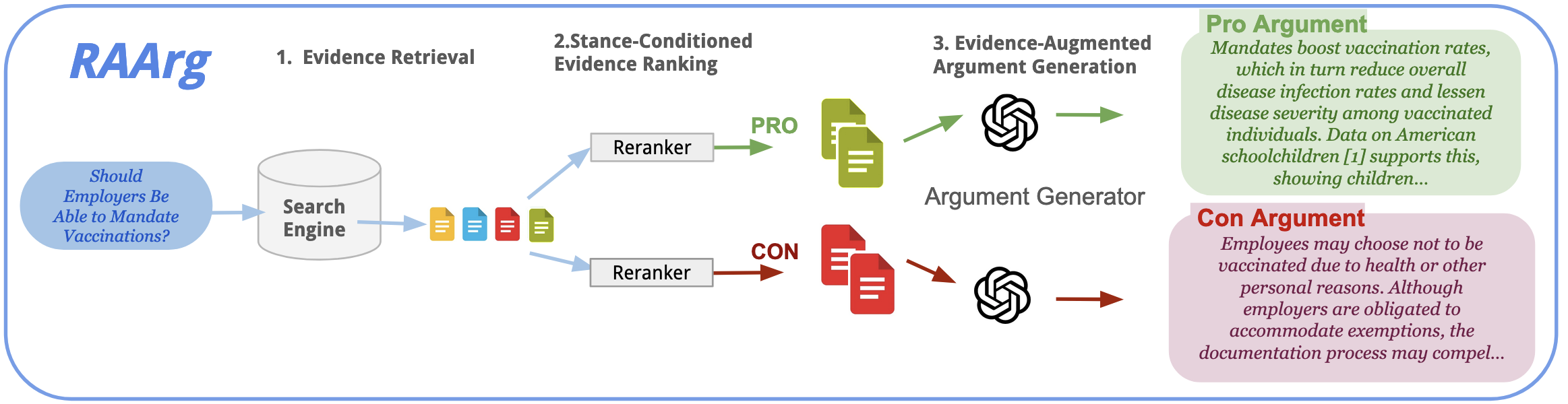}
        \caption{A high level overview of the task of Retrieval Augmented Argumentation (RAArg) generating pro and con arguments conditioned on retrieved (and reranked) evidence for a given controversial topic on ``vaccination''.}
        \label{fig:subfig1}
    \end{subfigure}
    \hfill
    \begin{subfigure}{\textwidth}
        \includegraphics[width=\linewidth]{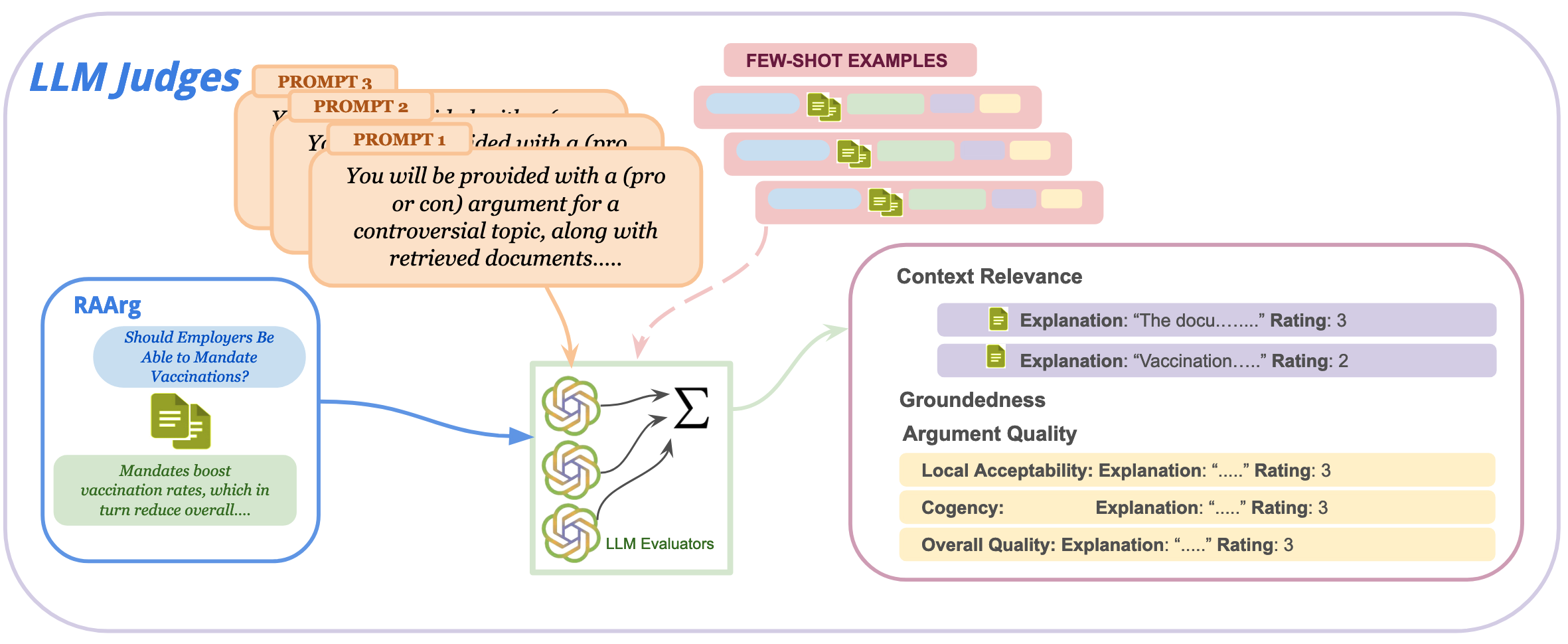}
        \caption{LLM Judges for computing fine-grained RAArg metrics to evaluate the retrieved (and reranked) evidence and the generated argument for a controversial topic (shown for the pro stance). }
        \label{fig:subfig2}
    \end{subfigure}
    \caption{Retrieval Augmented Argumentation (RAArg) and LLM Judges used for evaluation of RAArg.}
    \label{fig:fullprocess}
\end{figure*}
Computational argumentation~\cite{wachsmuth-etal-2017-computational,el-baff-etal-2019-computational}, or generating arguments for controversial topics—such as debates over banning abortion or the legalization of marijuana—inherently involve sensitive discussions and diverse opinions on the appropriate course of action. Well-informed argumentation requires retrieving publicly available sources of knowledge, such as blogs and news articles, which are often lengthy and noisy, and using them to create well-grounded and persuasive arguments. Large Language Models (LLMs) employed in a Retrieval Augmented Generation approach are a natural choice for this task due to their ability to handle large context lengths and strong performance across various tasks. To emphasize the importance of retrieved evidence for effective argumentation we call this task Retrieval Augmented Argumentation, or {\bf RAArg}, which we investigate in this paper.

While computational argumentation is a challenging task, even evaluating RAArg systems is a daunting challenge in itself. The generated arguments are often long and complex, making human evaluation costly and time-consuming. Thus, there is an increasing need for automated evaluation methods that can rigorously assess both the influence of retrieval involved in argument generation and the overall quality of the generated arguments, especially when dealing with a diverse set of controversial topics.

While previous research on automated RAG evaluation has largely focused on metrics like context relevance, answer relevance, and answer groundedness, often using LLM-based evaluation methods, also known as~\textit{LLM-as-a-Judge}~\cite{truera_rag_triad,saad-falcon-etal-2024-ares}. In contrast, work in argumentation has concentrated on assessing argument quality without considering the impact of retrieval~\cite{el-baff-etal-2019-computational,
wachsmuth-etal-2018-argumentation,
alshomary2020extractive}. Besides, these evaluations have notable limitations, particularly their reliance on single-score outputs that lack interpretability and provide minimal insights into specific performance dimensions like argument cogency, reasonableness, and performance at the granularity of individual documents. Additionally, many studies focus on short contexts~\cite{10.1145/3626772.3657957}, often chunking documents into passage-level contexts, which can omit essential contextual information. This lack of generalizability is problematic, especially for tasks like argumentation, which require evaluation of long, documents from the web, often unfamiliar to LLMs~\cite{answerAI2024}.

Specifically, a successful automated evaluation of an argumentation system should test the ability to~\textit{retrieve relevant context} and to~\textit{generate grounded}, and~\textit{high-quality arguments} for a diverse set of controversial topics. Additionally, the evaluation of the argument must agree with human preferences and provide cues of irrelevant retrieval and hallucination, ensuring that the system's performance is judged on, both the correctness of its answers and the influence of retrieved evidence. 

To address these challenges, we propose a systematic evaluation of LLM-based argumentation in a retrieval-augmented setting by validating multiple fine-grained metrics, which we call {\bf LLM Judges}, as different variants of LLM-based evaluation are needed for each of the metrics. The high level overviews of the task and our LLM Judges are depicted in~\cref{fig:fullprocess}.~\cref{fig:subfig1} shows the processing of our~\Task{} pipeline for a sample controversial topic and~\cref{fig:subfig2} shows our proposed fine-grained LLM Judges being employed to evaluate the retrieved evidence documents and the generated argument.

\begin{table*}[]
\resizebox{\textwidth}{!}{
\begin{tabular}{|c|cccccc|}
\hline
\textbf{Dataset} & \textbf{Task} & \textbf{\begin{tabular}[c]{@{}c@{}}Source of \\ Arguments\end{tabular}} & \textbf{Source of Evidence} & \textbf{\begin{tabular}[c]{@{}c@{}}Length of\\ Arguments\end{tabular}} & \textbf{\begin{tabular}[c]{@{}c@{}}Grounding \\ of Arguments\end{tabular}} & \textbf{\begin{tabular}[c]{@{}c@{}}Stance Conditioned\\ Arguments\end{tabular}} \\ \hline
\begin{tabular}[c]{@{}c@{}}ArguAna Counterargs\\ \cite{wachsmuth-etal-2018-retrieval}\end{tabular} & \begin{tabular}[c]{@{}c@{}}Counterargument\\ Retrieval\end{tabular} &  idebate.org & - & Short & X & \textbf{\chk} \\ \hline
\begin{tabular}[c]{@{}c@{}}Touche Argument Retrieval\\ \cite{bondarenko2020overview}\end{tabular} & Argument Retrieval & { \begin{tabular}[c]{@{}c@{}}ebatewise.org, \\ idebate.org, etc.\end{tabular}} & \begin{tabular}[c]{@{}c@{}}Arguments are assumed \\ to be directly available\end{tabular} & \multicolumn{1}{l}{} & X & \textbf{\chk} \\\hline
\begin{tabular}[c]{@{}c@{}}ChangeMyView\\ \cite{hua-wang-2018-neural}\end{tabular} & \begin{tabular}[c]{@{}c@{}}Counterargument\\ Generation (Dialog)\end{tabular} & {\begin{tabular}[c]{@{}c@{}}Reddit\\ ChangeMyView\end{tabular}} & - & Short & X & N/A \\ \hline
\begin{tabular}[c]{@{}c@{}}ArgTersely\\ \cite{winning,lin-etal-2023-argue}\end{tabular} & \begin{tabular}[c]{@{}c@{}}Counterargument\\ Generation\end{tabular} & { \begin{tabular}[c]{@{}c@{}}Reddit\\ ChangeMyView\end{tabular}} & - & Short & X & N/A \\ \hline
\begin{tabular}[c]{@{}c@{}}NPOV\\ \cite{chang-etal-2024-detecting}\end{tabular} & \begin{tabular}[c]{@{}c@{}}Neutral Point of\\ View Generation\end{tabular} & ProCon.org & - & Long & X & N/A \\ \hline
\begin{tabular}[c]{@{}c@{}}Dagstuhl-15512 ArgQuality\\ \cite{wachsmuth-etal-2017-computational, habernal-gurevych-2016-argument}\end{tabular} & Argument Generation & { \begin{tabular}[c]{@{}c@{}}createdebate.com,\\ procon.org\end{tabular}} & - & Mix & X & \textbf{\chk} \\ \hline
\textbf{\Dataset{}} & \textbf{\begin{tabular}[c]{@{}c@{}}Argument Retrieval\\ +Argument Generation\end{tabular}} &  \textbf{ProCon.org} & \textbf{\begin{tabular}[c]{@{}c@{}}Public Webpages\\Paired with Arguments\end{tabular}} & \textbf{Long} & \textbf{\chk} & \textbf{\chk} \\ \hline
\end{tabular}
}
\caption{Comparison of~\Dataset{} with other argumentation datasets.}
\label{tab:all_datasets}
\end{table*}

Specifically, our contributions are four-fold: (1) first, we demonstrate the feasibility of using~\textbf{model-based evaluation for the task of argumentation}; (2) we extend our analysis to the~\textbf{retrieval-augmented setting} (\Task{}), with evidence retrieved from Web documents; (3) we introduce a novel benchmark---\Dataset{}\footnote{ProCon webpage extraction code and documents will be released at~\url{https://github.com/emory-irlab/conqret-rag}}---\textbf{Con}troversial \textbf{Q}uestions for Argumentation and \textbf{Ret}rieval, consisting of human-authored arguments based on a popular debate portal; and (4) we investigate the performance of LLM Judges by introducing different types of errors into the arguments.

~\textit{To the best of our knowledge, this is the first work that investigates automated evaluation of retrieval augmented computational argumentation in a complex and realistic setting.}

In the rest of the paper, we first review related work to place our contributions in context (\cref{sec:relwork}). Then,~\cref{sec:task_des} presents a formal definition of the task of argumentation and a reference implementation using best RAG practices.~\cref{sec:dataset} details the construction process of the~\Dataset{} benchmark. In~\cref{sec:evaluation_methods}, we define multiple LLM Judges through novel and adapted variations of prompts, while~\cref{sec:results} validates them for both a prior argumentation dataset and over~\Dataset{}, demonstrating the feasibility of automated fine-grained evaluation of computational argumentation.

\section{Related Work}\label{sec:relwork}
We now describe some of the related work in argumentation and LLM evaluation.
\subsection{Computational Argumentation} The field of computational argumentation has focused on multiple problems in modeling the processes of debate and improving argumentation. Most approaches have attempted to model individual sub-tasks typically seen in debates like argument retrieval, argument construction, argument summarization, and counterargument generation. For instance,~\citet{hua-etal-2019-argument-generation} employed a BILSTM encoder-decoder neural network by retrieving evidence from Wikipedia, and showed improved topic relevance.~\citet{hua-wang-2018-neural} combined sophisticated retrieval mechanisms with a two-step generation model, enhancing the quality and appropriateness of generating counter-arguments. They evaluate their approaches over a counter-argument generation dataset based on the ChangeMyView website~\cite{changemyview}.
Building on the next utterance retrieval and generative dialogue systems,~\citet{le-etal-2018-dave} employed LSTMs to debate on controversial topics, showing promise on a corpus from the Convinceme website, an informal debating portal. Further,~\citet{el-baff-etal-2019-computational} select, arrange, and phrase Argument Discourse Units to synthesize arguments through pathos (emotional) and logos (logical) strategies for a dataset of 260 arguments and 10 debate topics~\cite{wachsmuth-etal-2018-argumentation}. 
~\citet{wachsmuth-etal-2018-retrieval} focus on retrieving potent counterarguments without prior topic knowledge, achieving significant accuracy through a model that evaluates argument similarities and differences. Lastly,~\citet{alshomary2020extractive} performed extracted snippets that more accurately represent the core reasoning of arguments collecting 73 arguments over 10 topics. These datasets have been summarised in~\cref{tab:all_datasets}.

\subsection{LLMs for Argumentation} In recent years, large language models (LLMs) have made significant progress across a variety of language tasks~\cite{srivastava2023beyond}, demonstrating particular promise in generating discourse-level text, which had previously been more challenging to model. Consequently, LLMs have also been employed to improve debate and argumentation~\cite{khan2024debating, alshomary-wachsmuth-2023-conclusion}.~\citet{li-etal-2023-halueval} create a manually annotated corpus of <topic, argument, counter-argument> triplets and then instruct tune a~\texttt{Llama-7B}~\cite{touvron2023llama} model to generate counterargument sentences.~\citet{verma2024auditing} evaluate LLMs' argument generation ability to incorporate style and external evidence from an automatically generated retrieval corpus.~\citet{mirza2024} perform a fine-grained evaluation of arguments from~\citet{wachsmuth-etal-2017-argumentation} in a pointwise fashion, by probing 15 argumentation metrics.

~\textit{In a nutshell, most of the approaches that focussed on both argument retrieval and argument generation were generally evaluated on informal debate corpora and generally lacked grounding over human-annotated evidence.} 

\subsection{LLM-based Evaluation of RAG Systems} Prior work on automated evaluations of RAG systems, particularly in multi-hop QA, has primarily focused on metrics like context relevance, answer relevance, answer groundedness~\cite{truera_rag_triad}. However, these metrics fall short when applied to tasks like argumentation, which require longer contexts and more detailed answers. In argumentation over controversial topics, the effects of irrelevant context and hallucinations are more pronounced, making evaluation crucial. Moreover, it is unclear if current evaluators can handle partial amounts of irrelevant context or provide actionable feedback. Moreover, the traditional reliance on single-score outputs~\cite{es-etal-2024-ragas, saad-falcon-etal-2024-ares} designed for short answers limits interpretability and usefulness in longer, more complex contexts such as arguments. This highlights the need for specialized evaluation methods that can better address the unique challenges of argumentation.

\section{Retrieval Augmented Argumentation}\label{sec:task_des}
In this section, we formally describe the task of retrieval augmented argumentation, or~\Task{}, and describe our reference implementation using state-of-the-art models and best practices.

\subsection{Task}
Given a controversial topic \( q \) and a stance \(s\), computational argumentation involves generating a stance-conditioned argument \( A_s \) with or without the retrieval of relevant evidence documents \( D_{s} \). 

\subsubsection{Evidence Retrieval} The first subtask, evidence retrieval, involves identifying relevant evidence documents $D_s$ to substantiate or refute the stance $s$ on a given topic $q$. 

\subsubsection{Argument Generation}
The second subtask, argument generation, focuses on using the retrieved documents to construct coherent and persuasive arguments, \( A_s \). This step requires the synthesis of the extracted evidence and ensuring that the arguments maintain logical consistency and are effectively communicated to address the controversial topic \( q \).

\subsection{Reference RAArg Implementation}
Our reference implementation follows the~\textit{retrieve-and-read} paradigm~\cite{lewis2020retrieval,izacard-grave-2021-leveraging}, where a retrieval system's top-ranked documents are input to a subsequent generator. Our~\Task{} system consists of: i) a 2-stage stance-conditioned evidence retriever and ii) a structured argument generator. Both components are evaluated with multiple strategies.
Note that~\Task{} is solely a reference implementation of a RAG system, designed to follow current best practices for retrieval and generation. Our aim is not to present a novel RAG system; rather, to implement a reasonable reference system following current best practices. Future work could explore more sophisticated retrieval and generation approaches.

For obtaining evidence relevant to the topic, we first retrieve the~\textit{top-k} relevant evidence documents $D_s$ for a given controversial topic $q$ and stance $s \in \{pro, con\}$ using the following ranking methodology.\\

\subsubsection{BM25 + LLM Reranking} Specifically, we first employ a two-stage BM25+Listwise Reranking paradigm popularised in studies like RankGPT~\cite{sun-etal-2023-chatgpt} and RankVicuna~\cite{pradeep2023rankvicuna}. This paradigm has shown SoTA results on a range of ranking tasks~\cite{thakur2021beir}. We use the name of the controversial topic as the query and index all documents with PyTerrier~\cite{macdonald2021pyterrier}. For LLM reranking, we use the~\texttt{Pyterrier\_GenRank}~\cite{Dhole_PyTerrier_Genrank} plugin with~\texttt{GPT-4o-mini}~\cite{openai_gpt4o_mini} as our reranker. We use two types of instructions--generic~\cite{sun-etal-2023-chatgpt} and stance-specific--along with a query and a list of documents, to generate a ranked list of document IDs for reranking. 
Other evidence retrieval and reranking methods could be investigated in the future, but as we will show this version performs adequately to enable generation of human-quality arguments.

\subsubsection{Few-Shot Argument Generation} We employ~\texttt{GPT-4o-mini} in a few-shot fashion, using the prompt reported in Appendix~\cref{fig:pro_argument_prompt_appendix}. For in-context examples, we select three controversial topics from the training set, along with their documents and corresponding expert-written arguments. Our choice of~\texttt{GPT-4o-mini} is based on its SoTA performance across various tasks, its ability to handle contexts up to 128K tokens, and its affordable pricing. We prompt the model to generate a structured argument that consists of a set of conclusions, each followed by premises that justify the conclusion.

Due to limitations of the (large) context size of popular LLMs, it is advantageous to prioritize documents that support the expected stance, as these are likely to produce a favorable argument. While RankGPT's instructions are useful, they remain generic and do not differentiate between documents supporting or opposing the stance. In contrast, we introduce stance-specific instructions during reranking, as illustrated in Appendix~\cref{fig:arg-pro-reranker}, enabling the generation of arguments tailored to each specific stance.

Furthermore, web documents could range in length from a few hundred tokens to as many as 100K tokens. Therefore, a context of 10 documents concatenated along with 3-few shot examples could exceed the (current) 128K token limit. In such cases, instead of trimming the documents only from the end, which could potentially exclude entire documents, we proportionally trim an equal portion from each document to minimize the possibility of removing a document in its entirety.

\section{The~\Dataset{} Benchmark}\label{sec:dataset}
In order to enable research on complex, evidence-based argumentation, we develop and present a novel benchmark,~\Dataset{}. The benchmark is constructed from a popular website of expert-generated arguments, which we further augment by retrieving the actual sources selected by the experts to be used as the ground truth evidence for retrieval augmented argument argumentation. The details of the dataset construction follow. 


\subsection{Web Scraping of ProCon.Org Pages}
We use ProCon.org, a debate portal featuring controversial questions with human-written pro and con arguments grounded in sources like journals and news outlets. Each argument includes paragraphs linked to external websites. We scrape these to gather grounded arguments on various topics, with details provided in~\cref{sec:web_scraping_appendix}.

\subsection{Retrieving Evidence Text from the Web}
\begin{figure}
    \centering
    \resizebox{1.1\columnwidth}{!}{
    \includegraphics[width=\textwidth]{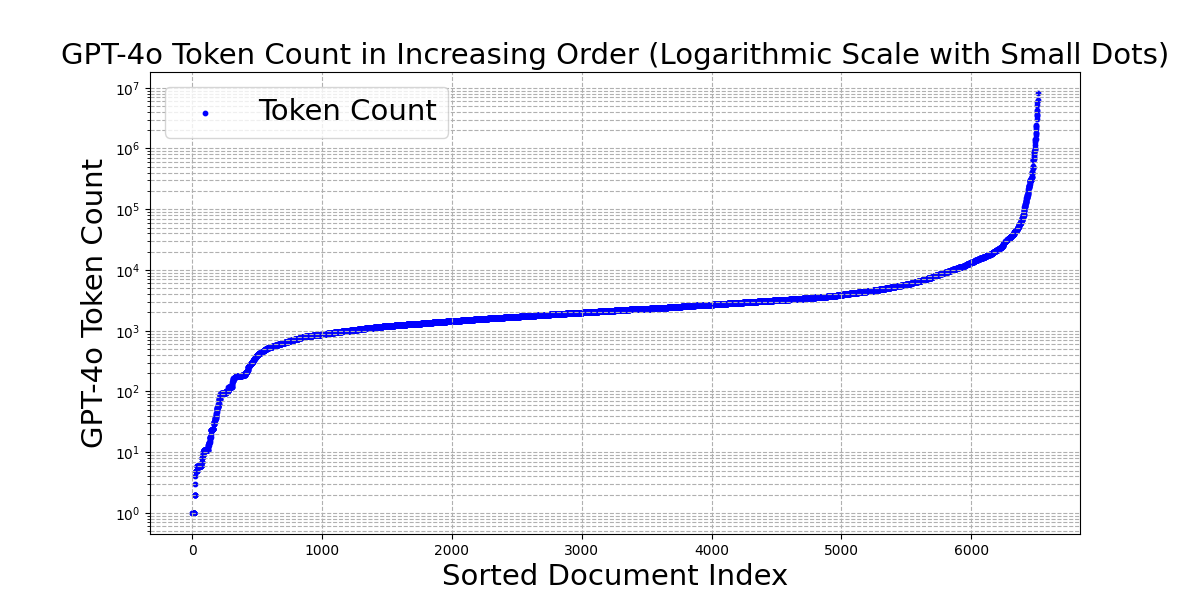}}
    \caption{Evidence Document Length Distribution. Most of the documents have 100 to around 10k tokens.}
    \label{fig:doc-length-distribution}
\end{figure}

Of the 7,177 sources cited on ProCon webpages, 6,514 (92\%) were successfully identified and extracted using Google or Bing. The remaining sources were either behind firewalls or required specialized parsing techniques, detailed in~\cref{sec:evidence_text_appendix}. Most source documents fall between a few hundred and 10,000 tokens, while the remainder are in the long tail, as illustrated by the document length distribution in~\cref{fig:doc-length-distribution}. 

\subsection{Construction of Query-Relevance Pairs}
To enable evidence retrieval evaluation, we construct query-document relevance pairs as follows:
\begin{enumerate}
    \item\textbf{Relevant Documents}: We use the scraped texts of each webpage mentioned in the~\texttt{sources} section of each topic to create pairs of queries and relevant documents.
    \item\textbf{Irrelevant Documents}: To create irrelevant (query, document) pairs, we randomly pick an equal number of documents from other queries.
\end{enumerate}

The pairs are split topicwise into two---70\% for train and 30\% for evaluation. Both sets do not have any topic or document overlaps shown in~\cref{conqret_summary}. This provides us with a set of 6.5k documents, and 98 topics.

\begin{table}[h!]
\centering
\resizebox{0.9\columnwidth}{!}{
\begin{tabular}{@{}lcc@{}}
\toprule
\hline
Statistic & Train & Test \\ \hline \midrule
Total topics & 68 & 30 \\
Avg. docs per topic & 69 & 64 \\
Avg. relevant docs per topic & 34 & 32 \\
Avg. docs per stance & 17 & 16 \\
Total evidence Documents& \multicolumn{2}{c}{6,500}\\
\hline
\bottomrule
\end{tabular}}
\caption{Summary of~\Dataset{} dataset statistics.
\label{conqret_summary}}
\end{table}

\begin{table*}[!ht]
\resizebox{\textwidth}{!}{
\begin{tabular}{|c|c|c|}
\cline{1-3}
\textbf{Format} & \multicolumn{1}{c|}{\textbf{Description}}  & \textbf{Output Computation}\\ \cline{1-3}
Direct~\cite{truera_rag_triad} & \begin{tabular}[c]{@{}c@{}}generate context relevance/preference/groundedness based on\\ evidence coverage, reasoning depth, coherence, and structure.\end{tabular} & Preference\\ \cline{1-3} 
Static-Rubric & \begin{tabular}[c]{@{}c@{}}evaluate individual rubrics like evidence coverage and reasoning\\
depth, selecting the most frequent preference.\\\end{tabular} & \begin{tabular}[c]{@{}c@{}}Preference (from aggregation \\ of preferences of static rubrics)\end{tabular} \\ \cline{1-3} 
\begin{tabular}[c]{@{}c@{}}G-Eval (Our adaptation of \\
~\citet{liu-etal-2023-g})\end{tabular} & \begin{tabular}[c]{@{}c@{}}generate context relevance/preference/groundedness via\\ static evaluation steps like G-eval\end{tabular} &  Preference\\ \cline{1-3}
\begin{tabular}[c]{@{}c@{}}Query-Rubric (Our adaptation of \\
~\citet{hashemi-etal-2024-llm})\end{tabular}& \begin{tabular}[c]{@{}c@{}}evaluate 20 query-specific rubrics and\\ selects the average rating or most frequent preference\\ \end{tabular} & \begin{tabular}[c]{@{}c@{}}Preference (from aggregation of \\ preferences of query-related rubrics)\end{tabular}\\ \cline{1-3}
RAG-Rubric (Ours) & \begin{tabular}[c]{@{}c@{}}generate context relevance, answer groundedness, and\\ preference using a single query-specific rubric prompt\end{tabular} & \begin{tabular}[c]{@{}c@{}}RAG Triad +\\ Preference (from aggregation of \\ preferences of query-related rubrics)\end{tabular}  \\ \cline{1-3}
RAG-Direct (Ours) & \begin{tabular}[c]{@{}c@{}}generate context relevance, answer groundedness, and preference\\directly through a single prompt\end{tabular} & \begin{tabular}[c]{@{}c@{}}RAG Triad +\\ Preference\end{tabular} \\ \cline{1-3}
RAG-Fine-Grained (Ours) & \begin{tabular}[c]{@{}c@{}}generate documentwise context relevance, answer groundedness,\\and preference directly via a single prompt.\end{tabular} & \begin{tabular}[c]{@{}c@{}}RAG Triad +\\ Preference\end{tabular} \\ \cline{1-3}
\end{tabular}
}
\caption{The different LLM-Judge prompts used to compute context relevance, groundedness, or preference against an expert written argument. The first four prompt variations use a separate prompt for each of the three metrics while the RAG-* variations are tasked to generate all the 3 evaluations through a single prompt.}
\label{tab:preferencemetrics}
\end{table*}

\section{Experiments and Analysis}\label{sec:evaluation_methods}
In this section, we describe the various LLM Judges used in our analysis. We first evaluate our LLM Judges in the non-RAG setting and then extend the evaluation to address the RAG counterpart.

\textbf{Datasets}: We utilize two datasets for our analysis---the \textbf{ArgQuality} Corpus~\cite{wachsmuth-etal-2017-argumentation} and \textbf{\Dataset{}} (Our new~\Task dataset introduced in Section~\ref{sec:dataset}). The~\textbf{ArgQuality} corpus contains 320 arguments, each annotated by experts across 15 quality dimensions using a 3-point Likert scale (low, medium, high).

\subsection{Fine-Grained Argument Quality Evaluation}
We now describe how we validate our LLM Judges for argument quality against fine-grained human annotations.

We assess argument quality through the 15 fine-grained dimensions proposed by~\citet{wachsmuth-etal-2017-computational}. Specifically, we introduce a~\textit{\textbf{Listwise}} evaluation prompt (Appendix~\cref{fig:argument_quality_prompt}), which presents the detailed definitions of these dimensions within a single inference call. These definitions, originally provided to argument expert annotators by~\citet{wachsmuth-etal-2017-computational}, are included along with the argument to be evaluated. 

The~\textit{\textbf{Listwise}} approach requires a single inference call and a reduced number of tokens compared to a~\textit{pointwise} approach, which would require 15 separate prompts--one for each dimension--resulting in a substantially higher token count. Due to the known lack of robustness of LLMs~\cite{liu-etal-2024-lost,dhole-etal-2023-nl}, we also explore the effect of varying the temperature and altering the order of the 15 dimensions in the prompt. Additionally, we employ a self-consistency~\cite{wang2023selfconsistency} style listwise prompt (\textit{\textbf{Listwise+SC}}) and compute the mean across three different orders. Our evaluations are compared against~\textit{pointwise} methods outlined in~\citet{mirza2024} and human crowd ratings from~\citet{wachsmuth-etal-2017-computational}.

For both datasets, expert annotator ratings are used, and the agreement between the model's predictions and the expert ratings is quantified using Krippendorff's $\alpha$~\cite{krippendorff2011computing}.

\subsection{RAG Evaluation}
\begin{table}[h!]
\centering
\resizebox{\columnwidth}{!}{
\begin{tabular}{llll}
\toprule
\hline
\textbf{QRELS} & \textbf{Ranker} & \textbf{P@10} & \textbf{NDCG@10} \\
\hline
\midrule
PRO & BM25 & .186 & .213 \\
    & BM25 + General Instruction & \textbf{.248}$^\alpha$ & \textbf{.312}$^\alpha$ \\
    & BM25 + Pro Conditioned Instruction & \textbf{.331}$^\alpha$ & \textbf{.434}$^\alpha$ \\
\hline
CON & BM25 & .245 & .262 \\
    & BM25 + General Instruction & .293$^\alpha$ &  .29\\
    & BM25 + Con Conditioned Instruction & .286 & .281 \\
\hline
PRO+CON & BM25 & .431 & .452 \\
    & BM25 + Pro Conditioned Instruction & \textbf{.538}$^\alpha$ & \textbf{.579}$^\alpha$ \\
    & BM25 + Con Conditioned Instruction & \textbf{.541}$^\alpha$ & \textbf{.582}$^\alpha$ \\
\hline
\bottomrule
\end{tabular}
}
\caption{Performance of Different Methods for Pro and Con Evidence Retrieval. The first column depicts which (query, document) relevance pairs were considered relevant for evaluation. $\alpha$ denotes significant improvements (paired t-test with Holm-Bonferroni correction~\cite{holm1979simple}, p < .05)}
\label{tab:ranking_models}
\end{table}

To comprehensively evaluate an argumentation system, it is essential to assess the quality of the final argument as well as the effect of retrieval. Particularly, we evaluate the following two metrics:

\begin{enumerate}
 \item \textbf{Context Relevance}---How relevant are the evidence documents to the controversial topic? 
\item \textbf{Argument Groundedness}---How grounded are the generated arguments in the retrieved evidence? 
\end{enumerate}

\subsubsection{Context Relevance: Traditional Eval}
To measure context relevance, we first use the human-annotated evidence in the form of qrels or (query, document) pairs to compute traditional IR metrics, viz., nDCG@k~\cite{DBLP:journals/tois/JarvelinK02} and P@k for~\Dataset{}. The results for the different retrievers are shown in~\cref{tab:ranking_models}.

In order to validate our LLM Judges, we resort to using these QRELS, similar to many automatic relevance labeling studies~\cite{rahmani2024llmjudge,10.1145/3539618.3592032, 10.1145/3626772.3657992}. 

We now describe our LLM Judges in the following subsection.
\subsubsection{Context Relevance: LLM Judges}
We introduce multiple LLM Judges for evaluating context relevance to simulate the more realistic setting when relevance annotations are unavailable. We particularly introduce different prompt variations, which take the topic $q$, documents $D$, and argument $A$ as inputs and generate context relevance scores along with argument groundedness and argument quality.

The different prompts are described below: 
\begin{table*}[t]
\resizebox{\textwidth}{!}{
\begin{tabular}{c|cccccccccccccc|cl}
\hline
\textbf{\begin{tabular}[c]{@{}c@{}}Argumentation\\ Quality\end{tabular}} & 
\rotatebox{70}{\textbf{Cogency}} & 
\rotatebox{70}{\textbf{Local Acceptability}} & 
\rotatebox{70}{\textbf{Local Relevance}} & 
\rotatebox{70}{\textbf{Local Sufficiency}} & 
\rotatebox{70}{\textbf{Effectiveness}} & 
\rotatebox{70}{\textbf{Credibility}} & 
\rotatebox{70}{\textbf{Emotional Appeal}} & 
\rotatebox{70}{\textbf{Clarity}} & 
\rotatebox{70}{\textbf{Appropriateness}} & 
\rotatebox{70}{\textbf{Arrangement}} & 
\rotatebox{70}{\textbf{Reasonableness}} & 
\rotatebox{70}{\textbf{Global Acceptability}} & 
\rotatebox{70}{\textbf{Global Relevance}} & 
\rotatebox{70}{\textbf{Global Sufficiency}} & 
\rotatebox{70}{\textbf{Average$^\alpha$}} \rotatebox{70}{\textbf{(Excluding OQ)}}  &\rotatebox{70}{\textbf{Overall Quality}} \\
\hline
\begin{tabular}[c]{@{}c@{}}Crowd / Expert\\ Wachsmuth et al. (2017)\end{tabular} & .27 & .49 & .42 & .18 & .13 & .41 & .45 & .42 & .54 & .53 & .33 & .54 & .44 & -.17 & .36  &.43 \\ \hline
\begin{tabular}[c]{@{}c@{}}LLMs Pointwise / Expert\\ Mirzakhmedova et al. (2024) Palm2\end{tabular} & .13 & \textbf{.59} & .47 & .47 & .2 & .42 & .17 & \textbf{.47} & .53 & \textbf{.51} & .43 & .43 & .58 & .46 & .42  &.29 \\ \hline
\begin{tabular}[c]{@{}c@{}}LLMs Pointwise / Expert\\ Mirzakhmedova et al. (2024) GPT3.5\end{tabular} & -.15 & .43 & .41 & -.4 & -.22 & \textbf{.62} & .48 & .39 & \textbf{.57} & .43 & .21 & .42 & \textbf{.64} & -.27 & .25  &.02 \\ \hline
\begin{tabular}[c]{@{}c@{}}LLMs Listwise / Expert\\ (Ours) GPT3.5\end{tabular} & .13 & .19 & .22 & .05 & .15 & .13 & .25 & .28 & .36 & .07 & .2 & .3 & .18 & -.02 & .18  &.16 \\ 
\begin{tabular}[c]{@{}c@{}}LLMs Listwise + SC / Expert\\ (Ours) GPT3.5\end{tabular} & .33 & .44 & .35 & .28 & .36 & .3 & .31 & .42 & .47 & .26 & .36 & .4 & .35 & .16 & .34  &.35 \\ 
\begin{tabular}[c]{@{}c@{}}LLMs Listwise + SC/ Expert\\ (Ours) GPT 4o\end{tabular} & \textbf{.47} & .45 & \textbf{.52} & \textbf{.57} & \textbf{.47} & .54 & \textbf{.50} & .38 & .56 & .42 & \textbf{.47} & \textbf{.56} & .31 & \textbf{.53} & \textbf{.48}  &\textbf{.48} \\ \hline
\hline
\end{tabular}}
\caption{Inter-annotator agreement (Krippendorff’s $\alpha$) between human experts and LLM annotations for each fine-grained argument quality dimension over ArgQuality benchmark~\cite{wachsmuth-etal-2017-argumentation}. $\alpha$ -- Average Quality is computed by averaging the first 14 dimensions. SC -- Self-Consistency.}
\label{tab:argument_quality_public}
\end{table*}
\begin{itemize}
    \item~\textit{\textbf{Listwise+RAG}}: Here, we instruct the model to generate other retrieval-based metrics, viz., argument groundedness and context relevance along with the argument quality. The context relevance is computed for a concatenation of all documents.
    \item~\textit{\textbf{Listwise+RAG Fine-Grained}}: This is the fine-grained counterpart of the~\textit{Listwise+RAG} metric where the context relevance is computed at the granularity of each document. The corresponding prompt is shown in~\cref{prompt_rag_docwise}.
    \item~\textit{\textbf{Others}}: In addition to the above, we also explore other formats of prompts to compute context relevance. They are described in~\cref{tab:preferencemetrics,sec:contextrelevanceevaluationformats}. 
\end{itemize}

We use~\texttt{GPT-4o}~\cite{openai_gpt4o_mini} for performing the evaluations due to its SoTA performance for other tasks for longer contexts. We also analyze other models viz.,~\texttt{phi-3-small/medium-128k-instruct-4} (7B/14B params)~\cite{abdin2024phi}, which can ingest large contexts and find that they are unable to understand the intent of evaluation often generating related but undesirable content missing values or unformatted JSON. We find that~\texttt{GPT-4o-mini}~\cite{openai_gpt4o_mini},~\texttt{Llama-3.1-70B}~\cite{dubey2024llama} and~\texttt{Gemini-1.5-Flash}~\cite{reid2024gemini} are among the ones which show the necessary sensitivity to irrelevant context. Additional details on the varying levels of irrelevant context across different LLM Judges and models are provided in the Appendix~\cref{tab:modelwise_context_rel}.

\section{Results and Analysis}\label{sec:results}
We now present the results of each of the above evaluations on both the datasets. 

\subsection{Fine-Grained Argument Quality Evaluation}
We present the results of the agreement between human experts and various LLM judges in~\cref{tab:argument_quality_public}. We find that when we employ a listwise approach with GPT3.5, we obtain better overall quality than a previous pointwise implementation~\cite{mirza2024}, and employing a listwise approach with self-consistency further improves overall quality as well as most of the individual dimensions. We obtain further improvements when we employ GPT-4o in the listwise setting.

Specifically, our listwise LLM-judge exhibits the highest agreement with expert annotators for both overall argument quality and across various argumentation dimensions, even compared to human crowd workers. Furthermore, the overall agreement is aligned with the average across the different dimensions, indicating consistency and alignment of the overall quality annotation with the intrinsic dimensions of argument quality. Our listwise LLM Judge is able to achieve overall quality agreement on par with humans and better than the pointwise setting employed with Palm2~\cite{anil2023palm} and GPT3.5~\cite{mirza2024}. Self-consistency (SC) further improves the overall quality as well as the average quality.~\textbf{\textit{Our Listwise + SC}} is also better than~\citet{mirza2024}'s ``novice'' prompt variation for 3 out of 4 dimensions.

Further, we investigate if there is variation in argument quality prediction when we change the order of the 15 dimensions. We find that there is substantial variability when the order of the dimensions is changed. We show in Appendix~\cref{tab:individual_llm_results} (a)-(c) how the results of three different runs change. We hence use a combined mean rating for each dimension, and see that there is high agreement in both the overall quality and the average quality, only at the expense of 2 additional inference calls. We also find prompting in a zero-shot fashion is more effective as compared to few-shot (Detailed results are shown in Appendix~\cref{tab:individual_llm_results} and~\cref{fig:detailed_argquality_results}).
\begin{table}[]
\resizebox{\columnwidth}{!}{
\begin{tabular}{l|ccc}
\hline
\multicolumn{1}{l|}{\textbf{Preferences}} & \textbf{No RAG} & \textbf{+ RAG} & \textbf{+ RAG Fine-Grained}\\ \hline
\textbf{Listwise GPT-4o} & .184 & .395 &~\textbf{.428} \\
\textbf{Listwise+SC (Mean)} & .317 & .439 &~\textbf{.445} \\
\textbf{Listwise+SC (Majority)} & .335 & .428 &~\textbf{.461} \\
\hline
\end{tabular}
}
 \caption{Inter-annotator agreement (Krippendorff’s $\alpha$) between human annotators and individual LLM annotators for argument quality metric over~\Dataset{}.}
\label{tab:individual_llm_results_conquer}
\end{table}

For our dataset~\Dataset{}, we present the results of the human agreement with three prompting strategies in~\cref{tab:individual_llm_results_conquer}. We find that when the LLM Judges are prompted to generate the individual RAG metrics before providing judging argument quality, they demonstrate better argument quality prediction. Additionally, employing the RAG metrics in a fine-grained fashion to obtain document-level scores further improves argument quality prediction. This demonstrates the benefit of retrieval-based metrics like context relevance and groundedness in helping provide better estimates of argument quality. Further details are mentioned in~\cref{argumentation_quality_conqret}. We also show a sample generation and its evaluation in Appendix~\cref{fig:sample_generation}.

\begin{table}[]
\centering
\resizebox{\columnwidth}{!}{
\begin{tabular}{c|ccc}
\hline
\textbf{\% Irrelevant Context} &   \textbf{True Precision} &\textbf{\begin{tabular}[c]{@{}c@{}}LLM Prediction\\ Direct (Trulens)\end{tabular}} & \textbf{\begin{tabular}[c|]{@{}c@{}}LLM Predicted Precision \\ (Listwise+RAG Fine-Grained)\end{tabular}}  \\
\multicolumn{1}{l|}{} &    && \multicolumn{1}{l}{} \\
\hline
0 &   .505 &.924 & .548 (+.043) \\
10 &   .344 &.845 & .467 (+.123) \\
20 &   .248 &.821 & .353 (+.106) \\
50 &   .097 &.907 & .171 (+.074) \\
70 &   .038 &.817 & .059 (+.021) \\
\hline
\end{tabular}}
\caption{Comparing the True Precision with the Precision of the Listwise+RAG Fine-Grained Approach. Values in brackets indicate absolute errors.}
\label{tab:precision_comparison}
\end{table}
\subsection{How close are context relevance predictions to Human Annotated Qrels?}
To validate how well our best LLM-Judge viz.~\textit{Listwise + RAG Fine-Grained} predict context relevance, we compare its precision at different levels of irrelevant content with the true precision, obtained through human annotated qrels. We evaluate how sensitive are context relevance predictions across different levels of irrelevant context. We specifically evaluate the context relevance by replacing some percentage of the retrieved documents with random irrelevant documents. We measure for no (0\%), low (10\%, 20\%), and high (50\%, 70\%) amounts of irrelevant content.

First, we compare it with the most popular approach of computing context relevance directly i.e. the direct metric~\cite{truera_rag_triad}. As we see in~\cref{tab:precision_comparison}, the Direct metric does not strictly decrease with increasing levels of irrelevant context lacking the necessary context sensitivity. However, our Listwise+RAG Fine-grained approach decreases gradually with increasing irrelevant context. Besides, our fine-grained approach provides precision estimates close to the True Precision. Also note that the Direct metric does not provide precision estimates at the granularity of individual documents, unlike the Listwise+RAG Fine-grained approach.

We also find that most fine-grained metrics, as shown in Appendix~\cref{tab:context_rel,tab:modelwise_context_rel} show better degradation with increasing irrelevant context than single-score metrics. Appendix~\cref{fig:context_relevannce_sensitivity_front} and~\cref{tab:context_rel} show the results when evaluated with GPT-4o. We find that while most metrics decrease with increasing irrelevant context, the decrease is not fully monotonic. The RAG-Direct and RAG-Rubric metrics display a near-perfect monotonic decrease (Pearson $\rho \approx$ -1), while the direct metric assigns the same context relevance score to 10\% and 50\% of irrelevant content. This trend is consistent across arguments of both stances. This demonstrates that single score metrics are less useful in practice vis-à-vis fine-grained metrics, like~\textit{Listwise+RAG Fine-grained} which are consistently sensitive to irrelevant context as well as interpretable.

\subsection{Do argument groundedness predictions reflect the number of hallucinations appearing in the argument?}
\begin{figure}[ht]
    \centering
\textit{``\textcolor{brown}{Despite at least six deaths from the laundry pod challenge, TikTok 
\sout{persists in promoting dangerous challenges from daring people to shave down their teeth with nail files}} 
\textcolor{purple}{has not seen any fatalities from its challenges, and the platform actively promotes safe practices among users}...\textcolor{brown}{[1]}''}
\caption{Hallucinated sentences (in purple) inserted into the argument by converting grounded sentences into sentences contradictory to the grounded evidence}
\label{fig_perturb}
\end{figure}
In the context of controversial topics, hallucination becomes increasingly significant. When assessing argumentation, it is essential to evaluate to what extent LLM judges are reliable for identifying hallucinated content in their argument.
\begin{figure*}[t]
    \centering
    \includegraphics[width=\textwidth]{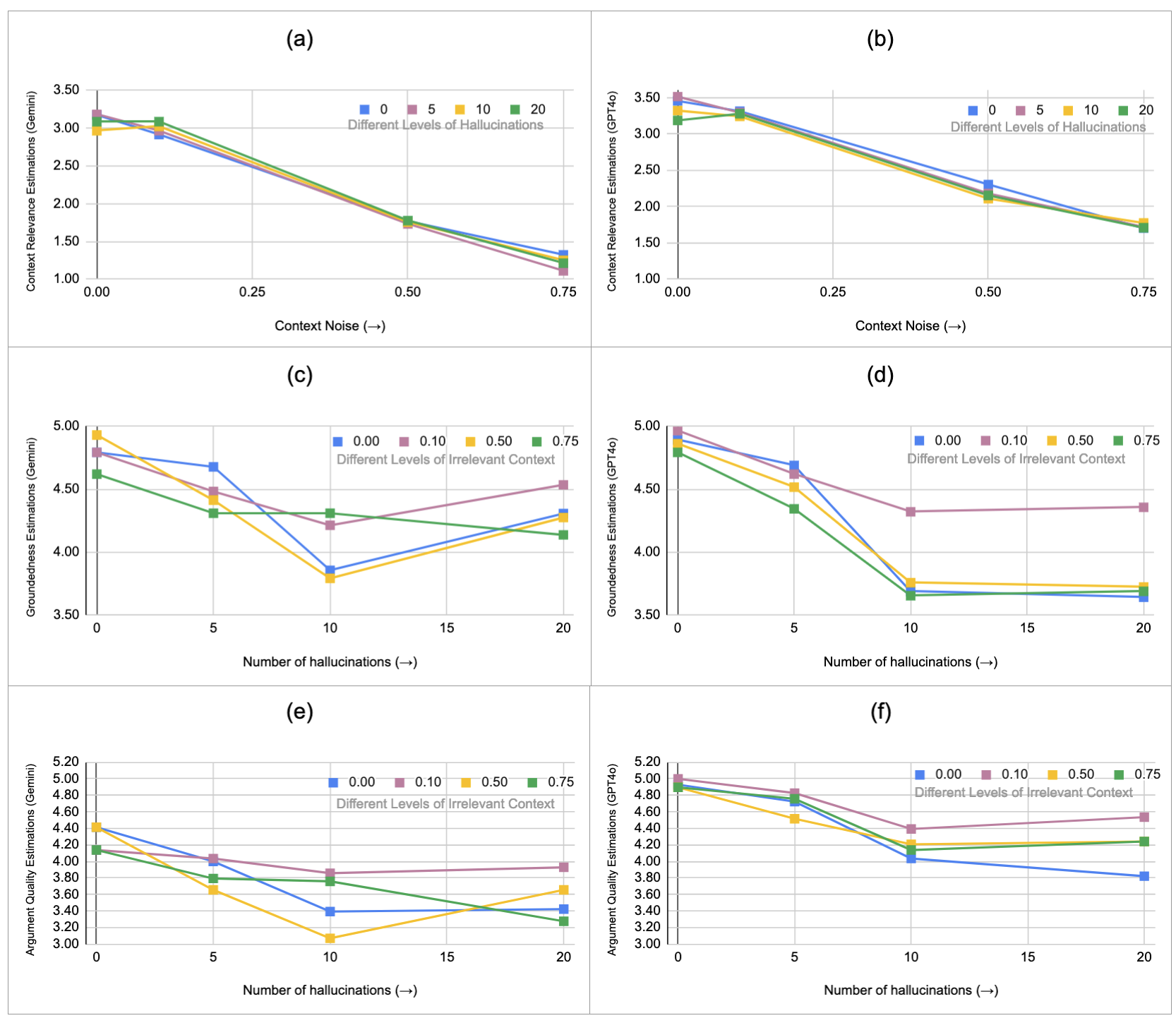}
    \caption{Metric consistency analysis for gemini-1.5-flash (left) and GPT-4o (right): Comparing the influence of both -- increasing irrelevant content and increasing hallucinations -- to see their effects on the 3 metrics.}
    \label{fig:intra-metric-comparison}
\end{figure*}

To simulate varying levels of hallucinations, we start with expert-written arguments paired with their corresponding documents from~\Dataset{}, then select a random subset of sentences from the original argument for modification. These selected sentences are replaced with hallucinated sentences, resulting in a modified argument as shown in~\cref{fig_perturb}. We anticipate that an effective metric will show a gradual reduction in its score as the proportion of modified sentences increases. To generate the hallucinations, we prompt~\texttt{GPT-4o} with the original argument and the original documents, and instruct the model to replace already grounded sentences by generating sentences which are contradictory to the original sentence as well as the content appearing in the document. We generate such modified arguments by modifying 5, and 20 sentences. The complete prompt is shown in Appendix~\cref{fig:modified_argument_prompt}. We find that all metrics are effective in displaying explainable differences across varying levels of groundedness. The results of the same are shown in Appendix~\cref{tab:answer_groundedness}.

\subsection{Metric Consistency Analysis}
As our~\textit{Listwise+RAG Fine-Grained} LLM Judge generates multiple metrics, it is also imperative to understand how they affect each other. For instance, does irrelevant context affect the evaluation of groundedness and argument quality? 

To measure the same, we increase the levels of~\textbf{both}---\textit{irrelevant context} (0\%, 10\%, 50\%, 75\%) and~\textit{hallucinated sentences} (0, 5, 20)\footnote{To account for varying argument lengths, we specify the hallucinated content in terms of 0, 5, and 20 sentences which correspond to 0\%, 16\%, and 63\% of an average 32-sentence argument.}--systematically and evaluate the effects of our metrics using~\texttt{gemini-1.5-flash} and~\texttt{GPT-4o} as our evaluators. We present the results in~\cref{fig:intra-metric-comparison}.

We find that across these coarse-grained levels of hallucinated infusions, context relevance scores and their monotonic degradations remain unaffected (\cref{fig:intra-metric-comparison} (a) and (b)). Moreover, at varying coarse-grained levels of irrelevant context, the groundedness scores and argument quality decrease monotonically with increasing hallucinated content. When we further add more fine-grained levels of hallucinations namely (2, 10), we find that context relevance scores are not strongly informative. The scores decrease with increased levels of injected irrelevant context.  
However, they are unable to distinguish fine-grained levels of irrelevant context.~\texttt{GPT-4o} seems more reliable than~\texttt{Gemini} for groundedness estimation. 

Argument quality estimations follow a decreasing trend at some levels of irrelevant context.~\texttt{GPT-4o} argument evaluations are more resistant to context than~\texttt{Gemini}. For a low number of hallucinations, argument quality evaluations are more reliable.

In summary, through our metrics, we observe that LLMs can be reliably used for estimating context relevance. However, prompt-based LLM Judges are not able to reliably evaluate argument groundedness and overall quality. For all metrics, LLM judges are not able to distinguish reliably between finer levels of non-relevant evidence or hallucinated arguments.

These results have important implications for argumentation as well as language modeling methods relying on LLMs for eliciting pseudo-human preferences, or employing them as teachers for subsequent distillation, as our methods reveal several shortcomings. 

\section{Conclusions and Future Work}\label{sec:conclusion}
Human annotation of complex arguments and corresponding evidence requires extensive knowledge of the topics, and copious amounts of reading of evidence for each argument, making fine-grained annotations infeasible at scale.
This paper introduces an automatic evaluation procedure and provides its validation without the need for large-scale scale human-annotation. 
We propose the first study on using model-based evaluation for retrieval-augmented argumentation systems. This problem is more important as long documents and complex arguments are involved over controversial topics. We demonstrate how argumentation can be effectively evaluated over two datasets, showing that our proposed LLM judges are more efficient, interpretable, and align strongly with human evaluations. Additionally, we introduce a novel benchmark~\Dataset{} designed to evaluate both retrievers as well as RAG systems in scenarios that closely resemble real-world contexts, characterized by long documents. Our contribution also includes the proposal of new and modified metrics through LLM judges for measuring context relevance and answer groundedness, and we describe methods to validate these metrics without requiring large-scale annotation. We find that larger models exhibit necessary degradation on fine-grained metrics, in contrast to popularly employed metrics that provide single, often uninterpretable, scores.

Our work has several crucial applications. Our LLM judges, in addition to demonstrating high agreement with human evaluations, also showed the capability to compute multiple metrics in a single inference call. This encourages research on a holistic evaluation of retrieval-augmented systems, allowing metrics to complement and inform each other during the evaluation process. Furthermore, these fine-grained metrics are promising, as they offer multiple cues to improve the training and alignment of LLMs, ultimately leading to more accurate and reliable systems.

\section{Limitations}\label{sec:limitations}
Our study does have some limitations, primarily related to the choice of models used for evaluation, namely GPT-4o, GPT-4o-mini, LLaMA-3.2-70B, and Gemini-1.5-Flash. While these models are SoTA in most aspects, additional models might provide more comprehensive performance insights. However, the selected models are highly representative of current capabilities in language models, and the findings hence offer valuable perspectives. Challenges such as the quadratic growth in memory requirements with increasing context lengths and occasional difficulties in handling specific instructions with large contexts in LLaMA-3.1 did arise, but these issues are inherent to many models and do not detract from the overall results.

\section{Ethical Considerations}
Controversial topics often involve complex disagreements on political, financial, or life-threatening issues and hence any computational argumentation system, built with LLMs should be treated as a part of a broader socio-technical context, where LLMs act as subsystems~\cite{dhole-2023-large}. Moreover, these systems, functioning as black boxes, camouflage the relationship between their outputs and the extensive pre-training data they are based on, making it hard to deduce the rationale for an argument or an evaluation score with high certainty.

On the other hand, LLMs have demonstrated biases across political spectrums and exhibit disparities in fairness across different languages and cultures, rendering them unsuitable for making impartial decisions on sensitive matters. Furthermore, when deployed in real-world applications, these systems are vulnerable to adversarial exploitation. For instance, adversaries may inject irrelevant content into retrieval processes or prompt the models to generate biased or misleading arguments that unfairly favor specific positions, often without proper attribution.

Our work highlights the susceptibility of LLM-based evaluations, emphasizing both their robustness and sensitivity. This underscores the critical need for robust evaluation frameworks that can assess argument quality while detecting shifts in context to ensure fairness. Additionally, LLM judges may themselves exhibit biases towards particular arguments. For this reason, we advocate for fine-grained, interpretative evaluations, which we believe will prove relatively more effective in practice than traditional single-score metrics. Our research is intended to encourage effort in that direction.

\section*{Acknowledgements}
This material is based upon work supported by the U.S. Department of Homeland Security under Grant Award Number 17STQAC00001-07-04, NSF awards (SaTC-2241068, IIS-2339198), a Cisco Research Award, and a Microsoft Accelerate Foundation Models Research Award. The views and conclusions in this document are those of the authors and should not be interpreted as necessarily representing the official policies, either expressed or implied, of the U.S. Department of Homeland Security or the National Science Foundation.

\bibliography{mypaper}

\appendix

\section{Web Scraping Public Websites}\label{sec:web_scraping_appendix}
In this section, we describe how we web scraped the ProCon.org website to create~\Dataset{}.

Initially, we resorted to Python based web-scraping to automate the extraction of HTML links from all debate pages on the website. Given the uniform yet occasionally divergent structure of these HTML pages, we wrote specific rules to accommodate variations in HTML and PDF content presentation. 

The scraped content from each page was systematically organized into a JSON format. The structure of this JSON includes multiple keys:~\texttt{title},~\texttt{introduction}, a list of ~\texttt{pro} and~\texttt{con} arguments, and a~\texttt{sources} field. Each argument is further divided into~\texttt{title} and~\texttt{paragraphs}, where~\texttt{paragraphs} contain multiple text segments. The~\texttt{sources} field includes details such as the local ID, the global document ID, the source title, and, if available, the URL. Notably, many sources did not include a direct URL, necessitating retrieving them from the next step as described below.

\section{Retrieving Evidence Text from the Web}\label{sec:evidence_text_appendix}
For relevant evidence, we leveraged the~\texttt{sources} listed on each debate page. The~\texttt{source} field consists of a title and optionally a URL. Wherever direct URLs were absent, the source's title was used as a query in Google or Bing search engines to retrieve the corresponding URL. We primarily used Google Search to obtain the first search result and employed Bing Search as a secondary option. The content of these URLs was then further scraped using custom HTML parsing and PDF parsing code. We removed webpages that couldn't be extracted due to firewall restrictions, or failed parsing. 

For the search engines, we used the googlesearch-python\footnote{\url{https://pypi.org/project/googlesearch-python/}} library and Azure's Bing API respectively. For HTML and PDF scraping, the BeautifulSoup\footnote{\url{https://pypi.org/project/beautifulsoup4/}} and PyPDF2\footnote{\url{https://pypi.org/project/PyPDF2/}} libraries were used. 

\begin{figure*}[htbp]
    \centering
    \fbox{%
        \parbox{0.9\textwidth}{%
        \small
            \textbf{Topic}: Should the U.S. Government Provide Universal Health Care \\[1ex]    \textbf{Stance}: Pro \\[1ex]
            \textbf{Retrieved Documents}:
            \begin{itemize}
                \item "Universal healthcare as pandemic preparedness: The lives and costs that could have been saved during ..."
                \item "Health Reimagined | Blue Shield of California | Program overview ..."
                \item "Improving the prognosis of health care in the USA - PubMed ..."
                \item \ldots
            \end{itemize}
            \textbf{GPT-4o Mini Argument}:
            ``**Argument: The U.S. Government Should Provide Universal Health Care**\\**Conclusion 1: Universal health care would significantly improve public health outcomes.**  \\ The implementation of a universal health care system is associated with increased life expectancy and better health outcomes. A study published in *Frontiers in Pharmacology* found that universal health coverage (UHC) is positively correlated with life expectancy at birth (LEAB) and healthy life expectancy (HALE). Specifically, the study indicated that UHC could lead to significant increases in both metrics, with standardized coefficients showing UHC's strong influence on life expectancy (Ranabhat et al., 2018) [8]. Furthermore, during the COVID-19 pandemic, a single-payer system could have saved approximately 212,000 lives and avoided \$105.6 billion in medical expenses related to COVID-19 hospitaliz\\.....\\.....\\A universal health care system would ensure that all individuals have access to timely medical care, vaccinations, and preventive services, thereby improving the nation’s ability to respond to future public health crises. Countries with UHC were able to mobilize resources more effectively during the pandemic, demonstrating the importance of having a comprehensive health care system in place for public health emergencies (Gaffney et al., 2021) [4].\\In conclusion, the U.S. government should provide universal health care as it would lead to improved health outcomes, reduce financial burdens on families, offer economic savings, and enhance preparedness for future health crises. The evidence from both domestic and international studies supports the transition to a universal health care system as a necessary step toward a healthier, more equitable society.''\\
            -----------------------------------------------------------------------------------------------------------------------------------------
            \textbf{\\Evaluation Summary} \\[1ex]
            \textbf{Context Relevance of Retrieved Documents}:
            \begin{itemize}
                \item \textbf{Document 1}: Score 5 - Highly relevant; addresses the impact of universal healthcare on saving lives during the COVID-19 pandemic.
                \item \textbf{Document 2}: Score 3 - Provides broad insights but lacks specific focus on universal healthcare.
                \item \textbf{Document 3}: Score 4 - Discusses healthcare prognosis in the USA, including benefits of universal healthcare; moderately relevant.
                \item \ldots
            \end{itemize}
            \textbf{Argument Quality Ratings}:
            \begin{itemize}
                \item \textbf{Answer Relevance}: Rating 5 - The argument is highly relevant, directly addressing the topic with a clear pro stance and specific points.
                \item \textbf{Answer Groundedness}: Rating 5 - Accurately reflects the content of the provided documents, citing specific studies and data points.
                \item \textbf{Cogency}: Rating 5 - Presents acceptable, relevant, and sufficient premises supporting its conclusions.
                \item \textbf{Clarity}: Rating 5 - Clear and well-structured, with logical flow and understandable language.
                \item \ldots
            \end{itemize}
            \textbf{Overall Quality}: Rating 5 - The argument is excellent, with strong evidence, clear structure, and persuasive reasoning.
        }
    }
    \caption{GPT-4o-mini RAG example output (top) and its LLM-Judge evaluation (bottom).}
    \label{fig:sample_generation}
\end{figure*}

\begin{table*}[h]
\centering
\begin{tabular}{|>{\raggedright\arraybackslash}p{0.9\textwidth}|}
    \hline
    \\
    You need to generate a persuasive argument which has a "pro" stance towards a controversial topic (i.e. you stand "for" the topic). For eg, if the topic is "Is Light a particle", then you should argue why light is indeed a particle. You will be given a collection of documents that provide evidence of a "pro" stance. In your argument, you need to summarise each document in 2 to 4 sentences by focussing on the main takeaways of each document. The argument should focus on the provided topic and convey the specific technical details on how the given documents support the pro stance of the topic. The argument should be overall coherent, take a pro stand and be detailed. If a document is unrelated to the topic, ignore it. \\
    \colorbox{pink}{\textbf{TOPIC}}\\
    \colorbox{pink}{\textbf{EVIDENCE DOCUMENTS (PRO)}}\\ \\
    \hline
\end{tabular}
\caption{Prompt Used for Argument Generation}
\label{fig:arg-gen}
\end{table*}

\begin{table*}[h]
\centering
\begin{tabular}{|>{\raggedright\arraybackslash}p{0.9\textwidth}|}
    \hline
    \\
    \texttt{You are an intelligent assistant that ranks documents supporting the 'pro' position of a given controversial query, meaning documents that provide evidence in favor of the `pro' argument. For instance, if the controversial query is `Is light a particle?', your task is to retrieve and rank documents that provide evidence supporting the argument that `light is a particle.'} \\ 
\texttt{I will provide you with \{\colorbox{pink}{k}\} documents, each identified by a number. \newline Rank the documents based on their relevance to supporting the pro position for the controversial query: \{\colorbox{pink}{query}\}. \newline Document 1: \newline \textless \colorbox{pink}{...Document 1 text...}\textgreater \newline . \newline . \newline Document k: \newline \textless \colorbox{pink}{...Document k text...}\textgreater }\\ 
\texttt{Search Query: \{\colorbox{pink}{query}\}. \newline Rank the \{\colorbox{pink}{num}\} documents above. Rank the documents based on their relevance to the pro position for the search query. List the documents in descending order using their identifiers, with the most relevant documents supporting the `pro' position listed first. The output format should be [1] > [2], and so on. Only respond with the ranking results; do not provide any explanations or additional words.}\\
\hline
\end{tabular}
\caption{Prompt for reranking documents based on their relevance to supporting the `pro' argument.}
\label{fig:arg-pro-reranker}
\end{table*}

\begin{table*}[h]
\centering
\begin{tabular}{|>{\raggedright\arraybackslash}p{0.9\textwidth}|}
    \hline
    \\
    \texttt{Craft a persuasive argument that takes a ``pro'' stance on a given controversial topic (i.e., you are in favor of the topic). For example, if the topic is ``Is light a particle?'', you would argue in support of the idea that light is indeed a particle. Using the provided documents, construct an argument that integrates multiple points, seamlessly incorporating evidence, historical context, and direct quotes.}
    \\
    \texttt{Your argument should follow the format shown in the provided examples: each argument should consist of multiple conclusions, with each conclusion followed by a set of premises that justify the conclusion. When referencing information from the documents, include appropriate citation(s) of the relevant documents in the form of [1], [2], etc., at the end of the premise.}
    \\
    \texttt{Ensure your argument includes detailed reasoning, is well-supported by the documents, and maintains a nuanced narrative that is both rich in detail and complexity. If a document is unrelated to the argument, omit it. Focus on creating a persuasive and human-like argument.}
    \\
    \texttt{\newline\textless \colorbox{cyan!20}{(Topic, Context, Argument)}\textgreater
    \newline\textless\colorbox{cyan!20}{(Topic, Context, Argument)}\textgreater
    \newline.
    .
    .}
    \\
    \texttt{Topic: \colorbox{pink}{\{topic\}}}
    \\
    \texttt{Context:}
    \texttt{Document 1: \newline \textless \colorbox{pink}{...Document 1 text...}\textgreater \newline . \newline . \newline Document k: \newline \textless \colorbox{pink}{...Document k text...}\textgreater }\\
    \hline
\end{tabular}
\caption{Prompt for crafting a ``pro'' stance argument on a controversial topic.}
\label{fig:pro_argument_prompt_appendix}
\end{table*}

\begin{table*}[h]
\centering
\small
\begin{tabular}{|>{\raggedright\arraybackslash}p{\textwidth}|}
    \hline
    \\
    You will be provided with an argument for a controversial topic, and you will be provided 15 dimensions to annotate the argumentation quality. You need to read the argument and evaluate each of these 15 dimensions on a scale from 1 to 3, where:
    
    \quad 3 = High
    
    \quad 2 = Medium
    
    \quad 1 = Low.
    
    You should provide your output only as a json with the dimension as the key and the value which contains the explanation and the rating.
    
    The following are the dimensions later followed by the argument:

    1) \textbf{local\_acceptability}: How would you rate the acceptability of the premises of the author’s argument?
    
    \textit{Local Acceptability}: A premise of an argument should be seen as acceptable if it is worthy of being believed, i.e., if you rationally think it is true or if you see no reason for not believing that it may be true. 
    If you identify more than one premise in the comment, try to adequately weight the acceptability of each premise when judging about their “aggregate” acceptability—unless there are particular premises that dominate your view of the author’s argumentation.
    
    2) \textbf{local\_relevance}: How would you rate the relevance of the premises of the author’s argument?
    
    \textit{Local Relevance}: A premise of an argument should be seen as relevant if it contributes to the acceptance or rejection of the argument’s conclusion, i.e., if you think it is worthy of being considered as a reason, evidence, or similar regarding the conclusion. 
    If you identify more than one premise in the comment, try to adequately weight the relevance of each premise when judging about their “aggregate” relevance—unless there are particular premises that dominate your view of the author’s argumentation. You should be open to see a premise as relevant even if it does not match your own stance on the issue.
    
    3) \textbf{local\_sufficiency}: How would you rate the sufficiency of the premises of the author’s argument?
    
    \textit{Local Sufficiency}: The premises of an argument should be seen as sufficient if, together, they provide enough support to make it rational to draw the argument’s conclusion.\newline
    ...\newline
    ...\newline
    13) \textbf{global\_sufficiency}: How would you rate the global sufficiency of the author’s argumentation?
    
    \textit{Global Sufficiency}: An argumentation should be globally sufficient if it adequately rebuts counter-arguments to its conclusion that can be anticipated.
    
    14) \textbf{reasonableness}: How would you rate the reasonableness of the author’s argumentation?
    
    \textit{Reasonableness}: An argumentation should be reasonable if it contributes to resolving the issue in a sufficient and acceptable way to everyone from the expected target audience.
    
    15) \textbf{overall\_quality}: How would you rate the overall quality of the author’s argumentation?
    
    \textit{Overall Quality}: ... Try to judge about the overall quality based on all those of your ratings that you think influence the overall quality of the given argumentation. If there is anything not covered by these ratings that influences your view of the author’s argumentation, also take that into account.\\
    \hline
\end{tabular}
\caption{Prompt used for evaluating argument quality along 15 dimensions based on documentation from~\citet{wachsmuth-etal-2017-computational}.}
\label{fig:argument_quality_prompt}
\end{table*}

\begin{table*}[h]
\centering
\begin{tabular}{|>{\raggedright\arraybackslash}p{0.9\textwidth}|}
    \hline
    \\
    \texttt{Craft a persuasive argument that takes a ``con'' stance on a controversial topic (i.e., you oppose the topic). For example, if the topic is ``Is light a particle?'', you would argue why light is not a particle. Using the provided documents, construct an argument that integrates multiple points, seamlessly incorporating evidence, historical context, and direct quotes.}
    \\
    \texttt{Your argument should follow the format shown in the provided examples: each argument should consist of multiple conclusions, with each conclusion followed by a set of premises that justify the conclusion. When referencing information from the documents, include appropriate citation(s) of the relevant documents in the form of [1], [2], etc., at the end of the premise.}
    \\
    \texttt{Ensure your argument includes detailed reasoning, is well-supported by the documents, and maintains a nuanced narrative that is both rich in detail and complexity. If a document is unrelated to the argument, omit it. Focus on creating a persuasive and human-like argument.}
    \texttt{\newline\textless \colorbox{cyan!20}{(Topic, Context, Argument)}\textgreater
    \newline\textless\colorbox{cyan!20}{(Topic, Context, Argument)}\textgreater
    \newline.
    .
    .}
    \\
    \texttt{Topic: \colorbox{pink}{\{topic\}}}
    \\
    \texttt{Context:}
    \texttt{Document 1: \newline \textless \colorbox{pink}{...Document 1 text...}\textgreater \newline . \newline . \newline Document k: \newline \textless \colorbox{pink}{...Document k text...}\textgreater }\\
    \hline
\end{tabular}
\caption{Prompt for crafting a "con" stance argument on a controversial topic.}
\label{fig:con_argument_prompt_appendix}
\end{table*}

\begin{figure*}[h]
\centering
\small
\begin{tabular}{|>{\raggedright\arraybackslash}p{0.9\textwidth}|}
    \hline
    \\
    \texttt{You are a RELEVANCE grader, evaluating the relevance and quality of the given context of evidence DOCUMENTS in relation to the provided controversial TOPIC. Assess the DOCUMENTS based on the following criteria and provide a boolean score (True or False) for each. Additionally, provide a brief explanation (2-4 lines) of your analysis.}
    \begin{enumerate}
        \item \texttt{Direct Relevance to the Topic: Do the DOCUMENTS collectively address the core aspects of the controversial TOPIC?}
        \item \texttt{Breadth of Coverage: Do the DOCUMENTS provide context relevant to multiple parts or aspects of the TOPIC?}
        \item \texttt{Quality of Evidence: Is the information in the DOCUMENTS credible and supportive of arguments related to the TOPIC?}
        \item \texttt{Applicability to Argumentation: Are the DOCUMENTS helpful for constructing a well-rounded argument for or against the TOPIC?}
        \item \texttt{Consistency with Topic Relevance: Are the DOCUMENTS consistently relevant throughout, without significant divergence into unrelated areas?}
        \item \texttt{Noise and Unrelated Content: Is the presence of noisier or unrelated DOCUMENTS minimal? (Lower scores for higher noise levels)}
    \end{enumerate}
    \\
    \texttt{Output Format:}
    \\
    \texttt{Your output should be in the following JSON format only. Provide a brief explanation (2-4 lines) of your analysis:}
    \\
    \texttt{\{ "explanation": "Your explanation here", "scores": \{ "direct\_relevance": true\_or\_false, "breadth\_of\_coverage": true\_or\_false, "quality\_of\_evidence": true\_or\_false, "applicability\_to\_argumentation": true\_or\_false, "consistency\_with\_topic\_relevance": true\_or\_false, "noise\_and\_unrelated\_content": true\_or\_false \} \}}
    \\
    \texttt{You should evaluate strictly and carefully. Do not output anything else apart from the JSON.}
    \\
    \hline
\end{tabular}
\caption{Fine-grained formatted Context Relevance Prompt}\label{prompt_fine_grained_context_rel}
\end{figure*}

\begin{figure*}[h]
\centering\small
\begin{tabular}{|>{\raggedright\arraybackslash}p{0.9\textwidth}|}
    \hline
    \\
    \texttt{You will be given a set of documents retrieved for a controversial topic. Your task is to rate the documents on one metric. Please make sure you read and understand these instructions carefully.}
    \\
    \\
    \texttt{Evaluation Criteria:}
    \\
    \texttt{Context Relevance (1-5) - Do the DOCUMENTS collectively address the core aspects of the controversial TOPIC? Do the DOCUMENTS provide context relevant to multiple parts or aspects of the TOPIC? Is the information in the DOCUMENTS credible and supportive of arguments related to the TOPIC? Are the DOCUMENTS helpful for constructing a well-rounded argument for or against the TOPIC? Are the DOCUMENTS consistently relevant throughout, without significant divergence into unrelated areas? Is the presence of noisier or unrelated DOCUMENTS minimal?}
    \\
    \\
    \texttt{Evaluation Steps:}
    \\
    \texttt{Step 1: Understand the Topic}
    \begin{itemize}
        \item \texttt{Identify the controversial topic under discussion. It is important to know what the debate or controversy centers on.}
        \item \texttt{Look for specific sub-topics or core aspects of the topic that are crucial for understanding all sides of the argument.}
    \end{itemize}
    \\
    \texttt{Step 2: Read Through the Documents}
    \begin{itemize}
        \item \texttt{Read all the documents carefully, taking note of whether they collectively cover the different facets of the controversy.}
        \item \texttt{Focus on how well each document addresses the topic and whether it provides substantial, relevant information.}
    \end{itemize}
    \\
    \texttt{Step 3: Assess Core Relevance}
    \begin{itemize}
        \item \texttt{Determine if the documents address key aspects of the controversy. Are they focused on important, core issues, or do they drift into unrelated topics?}
        \item \texttt{Check for consistent relevance throughout each document, making sure they stay focused on the controversy rather than diverging into side issues.}
    \end{itemize}
    \\
    \texttt{Step 4: Check for Credibility}
    \begin{itemize}
        \item \texttt{Evaluate the credibility of the information in each document. Does it come from reputable sources? Is it accurate, or does it appear biased or lacking evidence?}
        \item \texttt{Look for the presence of facts, data, or expert opinions that support arguments related to the topic.}
    \end{itemize}
    \\
    \texttt{Step 5: Examine Argument Support}
    \begin{itemize}
        \item \texttt{See how helpful the documents are in constructing a well-rounded argument. Do they cover both sides or multiple perspectives?}
        \item \texttt{Make sure that the documents, when taken together, contribute to understanding the full picture of the controversy.}
    \end{itemize}
    \\
    \texttt{Step 6: Rate for Noise/Unrelated Content}
    \begin{itemize}
        \item \texttt{Identify if there are documents or portions of documents that provide irrelevant or noisy information. Are there sections that do not contribute to the topic or seem off-track?}
        \item \texttt{The fewer irrelevant sections, the higher the rating.}
    \end{itemize}
    \\
    \texttt{Step 7: Assign a Rating (1-5)}
    \begin{itemize}
        \item \texttt{After reviewing all the above criteria, assign a rating based on how well the documents collectively meet the following standards:}
        \begin{itemize}
            \item \texttt{1 (Very Low Relevance): The documents barely touch on the topic or are mostly irrelevant.}
            \item \texttt{2 (Low Relevance): Some relevant content, but much is off-topic or lacks depth.}
            \item \texttt{3 (Moderate Relevance): The documents cover the topic but lack consistency or completeness.}
            \item \texttt{4 (High Relevance): Most of the content is relevant, with only minor off-topic sections.}
            \item \texttt{5 (Very High Relevance): The documents are consistently relevant and fully address all key aspects of the controversy.}
        \end{itemize}
    \end{itemize}
    \\
    \texttt{You should only output the following JSON:}
    \\
    \texttt{\{ "score": your\_score \}}
    \\
    \hline
\end{tabular}
\caption{G-Eval formatted Context Relevance Prompt}
\label{prompt_g_eval}
\end{figure*}

\begin{figure*}[h]
\centering
\small
\begin{tabular}{|>{\raggedright\arraybackslash}p{0.9\textwidth}|}
    \hline
    \\
    \texttt{You are tasked with generating a list of 20 'nuggets' based solely on the given query and its stance. A nugget is a key piece of information or point that might support or refute the query's stance. Next, evaluate the given context of DOCUMENTS and assess how many of the generated nuggets can be inferred from these DOCUMENTS. For each nugget, assign a score between 1 and 5, depending on how well the DOCUMENTS cover that nugget, where 1 means poorly covered and 5 means thoroughly covered.}
    \\
    \\
    \texttt{Output the results in the following JSON format:}
    \\
    \texttt{\{ "nuggets": [ \{"the first nugget": score\}, \{"the second nugget": score\}, ... \{"the 20th nugget": score\} ] \}}
    \\
    \\
    \texttt{Ensure that the model evaluates strictly based on the provided context and does not generate any additional information beyond the given instructions.}
    \\
    \hline
\end{tabular}
\caption{Query Rubric Formatted Context Relevance Prompt}
\label{prompt_query_rubric}
\end{figure*}

\begin{figure*}[h]
\centering
\small
\begin{tabular}{|>{\raggedright\arraybackslash}p{0.9\textwidth}|}
    \hline
    \\
    \texttt{You are tasked with performing the following analysis based on a given query, its stance (Pro or Con), provided DOCUMENTS, and two arguments (Argument 1 and Argument 2). Your goal is to assess context relevance, answer relevance, answer groundedness, and argument preference evaluation.}
    \\
    \\
    \texttt{Instructions:}
    \\
    \texttt{Context Relevance:}
    \begin{itemize}
        \item \texttt{Generate a list of 20 nuggets based solely on the given query and its stance. A nugget is a key piece of information or point that might support or refute the query's stance.}
        \item \texttt{Evaluate how well each nugget is covered by the provided DOCUMENTS.}
        \item \texttt{Assign a score between 1 and 5 to each nugget, where:}
        \begin{itemize}
            \item \texttt{1 = Poorly covered by the DOCUMENTS.}
            \item \texttt{5 = Thoroughly covered by the DOCUMENTS.}
        \end{itemize}
    \end{itemize}
    \\
    \texttt{Answer Relevance:}
    \begin{itemize}
        \item \texttt{Using the same 20 nuggets from the previous step, evaluate how well each nugget is addressed by Argument 1 and Argument 2 separately.}
        \item \texttt{Assign a score between 1 and 5 for each nugget under each argument, where:}
        \begin{itemize}
            \item \texttt{1 = Nugget is minimally or not at all covered by the argument.}
            \item \texttt{5 = Nugget is fully covered by the argument.}
        \end{itemize}
    \end{itemize}
    \\
    \texttt{Answer Groundedness:}
    \begin{itemize}
        \item \texttt{Generate a new list of 20 key nuggets derived solely from the content of the provided DOCUMENTS.}
        \item \texttt{Evaluate how well each nugget is covered by Argument 1 and Argument 2 separately.}
        \item \texttt{Assign a score between 1 and 5 for each nugget under each argument, where:}
        \begin{itemize}
            \item \texttt{1 = Poor coverage in the argument.}
            \item \texttt{5 = Excellent coverage in the argument.}
        \end{itemize}
    \end{itemize}
    \\
    \texttt{Argument Preference Evaluation:}
    \begin{itemize}
        \item \texttt{Based on the evaluations from the previous steps, determine which argument best addresses each nugget.}
        \item \texttt{For each nugget, indicate your preference as:}
        \begin{itemize}
            \item \texttt{"Argument 1"}
            \item \texttt{"Argument 2"}
            \item \texttt{"Both"}
        \end{itemize}
    \end{itemize}
    \\
    \texttt{Output Format:}
    \\
    \texttt{Provide your results in a JSON object with the following structure:}
    \\
    \texttt{\{ "context\_relevance": \{ "nuggets": \{ "Renewable energy reduces carbon emissions": 5, "Initial cost of renewable energy is high": 4, "...": "...", "Renewable energy sources are unreliable": 2 \} \},}
    \\
    \texttt{"answer\_relevance": \{ "nuggets": \{ "Renewable energy reduces carbon emissions": \{ "Argument 1": 5, "Argument 2": 3 \}, "Initial cost of renewable energy is high": \{ "Argument 1": 2, "Argument 2": 5 \}, "...": "...", "Renewable energy sources are unreliable": \{ "Argument 1": 1, "Argument 2": 4 \} \} \},}
    \\
    \texttt{"answer\_groundedness": \{ "nuggets": \{ "Solar energy is abundant": \{ "Argument 1": 5, "Argument 2": 2 \}, "Wind energy depends on weather": \{ "Argument 1": 3, "Argument 2": 5 \}, "...": "...", "Hydroelectric power impacts aquatic life": \{ "Argument 1": 2, "Argument 2": 4 \} \} \},}
    \\
    \texttt{"argument\_preference\_evaluation": \{ "nuggets": \{ "Renewable energy reduces carbon emissions": "Argument 1", "Initial cost of renewable energy is high": "Argument 2", "...": "...", "Renewable energy sources are unreliable": "Both" \} \} \}}
    \\
    \texttt{Only provide your output in a valid JSON format and nothing else.}
    \\
    \hline
\end{tabular}
\caption{RAG  Rubric Formatted Prompt: The prompt is used to generate multiple scores including context relevance and groundedness}
\label{prompt_rag_rubric}
\end{figure*}

\begin{figure*}[h]
\centering\small
\begin{tabular}{|>{\raggedright\arraybackslash}p{0.9\textwidth}|}
    \hline
    \\
    \texttt{Task: You are tasked with evaluating two arguments on a controversial topic. One argument is written by a human, and the other is generated by a model, but you are not told which is which. Your evaluation should consist of four key metrics:}
    \\
    \\
    \texttt{Context Relevance:}
    \begin{itemize}
        \item \texttt{Evaluate the relevance of the provided evidence documents to the controversial topic for both arguments. Criteria:}
        \begin{itemize}
            \item \texttt{Evidence Alignment: Does the document contain information supporting or opposing the topic?}
            \item \texttt{Specificity: Is the content detailed in relation to the topic?}
            \item \texttt{Usefulness: Is the document useful for building an argument?}
        \end{itemize}
    \end{itemize}
    \\
    \texttt{Answer Relevance:}
    \begin{itemize}
        \item \texttt{Assess how relevant each argument is in addressing the controversial topic. Criteria:}
        \begin{itemize}
            \item \texttt{Topic Adherence: Does the argument directly address the topic with a clear stance (for or against)?}
            \item \texttt{Specificity: Is the argument specific and detailed in its response?}
            \item \texttt{Persuasiveness: Is the argument persuasive and comprehensive?}
        \end{itemize}
    \end{itemize}
    \\
    \texttt{Answer Groundedness:}
    \begin{itemize}
        \item \texttt{Evaluate how well each argument is grounded in the provided evidence documents. Criteria:}
        \begin{itemize}
            \item \texttt{Evidence Accuracy: Does the argument accurately reflect the content of the evidence documents?}
            \item \texttt{Evidence Citation: Does the argument cite specific parts of the evidence?}
            \item \texttt{Consistency: Is the argument consistent with the facts from the evidence?}
        \end{itemize}
    \end{itemize}
    \\
    \texttt{Argument Preference Evaluation:}
    \begin{itemize}
        \item \texttt{After evaluating the context relevance, answer relevance, and groundedness, determine which argument is more effective overall. Criteria:}
        \begin{itemize}
            \item \texttt{Coverage of Evidence: Which argument addresses key points and evidence better?}
            \item \texttt{Depth of Reasoning: Which argument offers more detailed insights?}
            \item \texttt{Coherence and Structure: Which argument is more logically structured and coherent?}
            \item \texttt{Stance Consistency: Which argument better adheres to the specified stance?}
        \end{itemize}
    \end{itemize}
    \\
    \texttt{Output Format:}
    \\
    \texttt{Your output should be in the following JSON format only. For each of the four metrics, provide a score (for context relevance, answer relevance, and answer groundedness) and a preference (for argument preference evaluation). Additionally, provide a brief explanation (2-4 lines) of your analysis for each score or preference.}
    \\
    \texttt{\{ "context\_relevance": \{ "explanation": "Your explanation here", "score\_argument\_1": your\_score, "score\_argument\_2": your\_score \},}
    \\
    \texttt{"answer\_relevance": \{ "explanation": "Your explanation here", "score\_argument\_1": your\_score, "score\_argument\_2": your\_score \},}
    \\
    \texttt{"answer\_groundedness": \{ "explanation": "Your explanation here", "score\_argument\_1": your\_score, "score\_argument\_2": your\_score \},}
    \\
    \texttt{"argument\_preference\_evaluation": \{ "explanation": "Your explanation here", "preference": "Argument 1" or "Argument 2" or "Tie" \} \}}
    \\
    \hline
\end{tabular}
\caption{RAG Direct formatted prompt used for generating multiple metrics including context relevance and answer groundedness}
\label{prompt_rag_direct}
\end{figure*}

\begin{figure*}[h]
\centering\small
\begin{tabular}{|>{\raggedright\arraybackslash}p{0.9\textwidth}|}
    \hline
    You will be provided with a (pro or con) argument for a controversial topic, along with retrieved documents. Your goal is to assess context relevance of each document, answer relevance, answer groundedness, and then 15 other argumentation metrics. For each of these metrics, you need to return an explanation and a rating. Your evaluation should consist of the following metrics:
    
    \textbf{context\_relevance (a list)}: Evaluate the relevance of each of the provided evidence documents to the controversial topic for the given argument.\\ Criteria:
    \begin{itemize}
        \item \textbf{Evidence Alignment}: Does the document contain information supporting or opposing the topic?
        \item \textbf{Specificity}: Is the content detailed in relation to the topic?
        \item \textbf{Usefulness}: Is the document useful for building an argument?
    \end{itemize}

    \textbf{Answer Relevance}: Assess how relevant each argument is in addressing the controversial topic.\\ Criteria:
    \begin{itemize}
        \item \textbf{Topic Adherence}: Does the argument directly address the topic with a clear stance (for or against)?
        \item \textbf{Specificity}: Is the argument specific and detailed in its response?
        \item \textbf{Persuasiveness}: Is the argument persuasive and comprehensive?
    \end{itemize}

    \textbf{Answer Groundedness}: Evaluate how well each argument is grounded in the provided evidence documents.\\Criteria:
    \begin{itemize}
        \item \textbf{Evidence Accuracy}: Does the argument accurately reflect the content of the evidence documents?
        \item \textbf{Evidence Citation}: Does the argument cite specific parts of the evidence?
        \item \textbf{Consistency}: Is the argument consistent with the facts from the evidence?
    \end{itemize}

    \textbf{Argument Quality}: After evaluating the context relevance, answer relevance, and groundedness, you need to annotate the argumentation quality based on 15 dimensions. You need to read the argument and evaluate each of these 15 dimensions on a scale from 0 to 5. You should provide your output only as a JSON with the dimension as the key and the value which contains the explanation and the rating. \\
    The following are the dimensions which are followed by the argument:

    1) local\_acceptance - ...\\
    ...\\
    15) overall\_quality - ...\\
    \hline
\end{tabular}
\caption{Listwise + RAG Fine-Grained formatted prompt used for generating multiple metrics including context relevance at the granularity of a document and 15 argumentation quality metrics}
\label{prompt_rag_docwise}
\end{figure*}

\begin{table*}[]
\resizebox{\textwidth}{!}{
\begin{tabular}{l|ccc|cccc}
\hline
\multicolumn{1}{c|}{\textbf{\begin{tabular}[c|]{@{}c@{}}Argumentation\\ Quality Dimension\end{tabular}}} 
&  {\begin{tabular}[c|]{@{}c@{}}Crowd\\(Wachsmuth\\et al.(2017))\end{tabular}} 
& {\begin{tabular}[c|]{@{}c@{}}Pointwise\\(Mirzakhmedova\\et al.(2024) GPT3.5)\end{tabular}} 
& {\begin{tabular}[c|]{@{}c@{}}Pointwise\\(Mirzakhmedova\\et al.(2024) Palm2)\end{tabular}}
& \begin{tabular}[c]{@{}c@{}}(a)\\Listwise
\end{tabular}& \begin{tabular}[c]{@{}c@{}}(b)\\Listwise
\end{tabular}& \begin{tabular}[c]{@{}c@{}}(c)\\Listwise
\end{tabular}& \begin{tabular}[c]{@{}c@{}}(a)+(b)+(c)\\Listwise+SC  
\end{tabular}\\ \hline
\multicolumn{1}{c|}{\textbf{Variable}} &   & & & & \begin{tabular}[c]{@{}c@{}} + Reordered  dimensions 1\end{tabular}& \begin{tabular}[c]{@{}c@{}} + Reordered  dimensions 2\end{tabular}& \\ \hline
\textbf{\textit{Average of dimension scores}}  &  .36&.25&.42& .31& .45& .44& \textbf{.48}\\ 
\textbf{Overall Quality Rating}  &  .43&.02&.29& .35& \textbf{.50}& .46& \textbf{.48}\\
\hline
\end{tabular}
}
\caption{Inter-annotator agreement (Krippendorff’s $\alpha$) between between experts and various LLM Judges (pointwise, listwise, listwise+SC) for the average quality metrics and average quality over ArgQuality.}
\label{tab:individual_llm_results}
\end{table*}

\begin{table*}[!ht]
    \centering
    \resizebox{\textwidth}{!}{
    \begin{tabular}{c|cccc|cccc}
    \hline
\multicolumn{1}{l}{} & 
\multicolumn{1}{l}{\begin{tabular}[c]{@{}c@{}}(a)\\ Listwise \\ Dimensions 1\end{tabular}}  & 
\multicolumn{1}{l}{\begin{tabular}[c]{@{}c@{}}(b)\\ + Reordered \\ Dimensions 1\end{tabular}} & \multicolumn{1}{l}{\begin{tabular}[c]{@{}c@{}}(c)\\ + Reordered \\ Dimensions 2\end{tabular}} & \multicolumn{1}{l|}{\begin{tabular}[c]{@{}c@{}}(a)+(b)+(c)\\+ Self Consistency\\\end{tabular}} & \multicolumn{1}{l}{\begin{tabular}[c]{@{}c@{}}(a)\\ Listwise \\ Dimensions 1\end{tabular}}  & 
\multicolumn{1}{l}{\begin{tabular}[c]{@{}c@{}}(b)\\ + Reordered \\ Dimensions 1\end{tabular}} & \multicolumn{1}{l}{\begin{tabular}[c]{@{}c@{}}(c)\\ + Reordered \\ Dimensions 2\end{tabular}} & \multicolumn{1}{l}{\begin{tabular}[c]{@{}c@{}}(a)+(b)+(c)\\+ Self Consistency\\\end{tabular}}\\
\hline
\textbf{\begin{tabular}[c]{@{}c@{}}Argumentation \\ Quality Dimensions\end{tabular}} & \multicolumn{4}{c|}{ZERO SHOT} & \multicolumn{4}{c}{5-SHOT} \\
\hline
\multicolumn{1}{l}{\textbf{Cogency}} & .369& .417 & .404 & \textbf{.471}& .388 & .333 & .418 & .435 \\
\textbf{Local Acceptability} & .321 & .425 & .434 & .455 & .474 & .466 & .466 & \textbf{.530}\\
\textbf{Local Relevance} & .283& .498 & .509 & \textbf{.523}& .287 & .511 & .338 & .438 \\
\textbf{Local Sufficiency} & .360& .431 & .478 & \textbf{.570}& .280 & .324 & .389 & .383 \\
\multicolumn{1}{l}{\textbf{Effectiveness}} & .287 & .447 & \textbf{.485}& .466 & .320 & .359 & .368 & .398 \\
\textbf{Credibility} & .320 & .449 & .481 & \textbf{.544}& .401 & .441 & .434 & .499 \\
\textbf{Emotional Appeal} & .447& .420 & .349 & .501 & .427 & .423 & .452 & \textbf{.515}\\
\textbf{Clarity} & .203 & .404 & .364 & .380 & .370 & .364 & .403 & .431 \\
\textbf{Appropriateness} & .366 & .542 & .562 & .561 & .550 & .564 & .559 & \textbf{.629}\\
\textbf{Arrangement} & .171 & \textbf{.467}& .423 & .423 & .329 & .399 & .342 & .420 \\
\multicolumn{1}{l}{\textbf{Reasonableness}} & .363 & .479 & .411 & .471 & .464 & .442 & .374 & \textbf{.486}\\
\textbf{Global Acceptability} & .382 & .519 & .480 & \textbf{.562}& .478 & .495 & .477 & .553 \\
\textbf{Global Relevance} & .193 & .387 & .297 & .315 & .228 & \textbf{.377}& .067 & .263 \\
\textbf{Global Sufficiency} & .287 & .351 & .435 & \textbf{.525}& .104 & .185 & .180 & .184 \\
\multicolumn{1}{l}{\textbf{Overall Quality}} & .350 & \textbf{.497}& .456 & .480& .417 & .385 & .386 & .446 \\
\multicolumn{1}{l}{\textbf{Average (excluding OQ)}} & .311 & .445& .437& \textbf{.483} & .364& .406 & .376 & .440 \\
\hline
\end{tabular}}
    \caption{Detailed results of argument quality across LLM Judges of different listwise orderings in both zero-shot and 5-shot setting for ArgQaulity.}
    \label{fig:detailed_argquality_results}
\end{table*}

\section{Argumentation Quality Evaluation over~\Dataset{}}\label{argumentation_quality_conqret}
For the~\textbf{\Dataset{}} corpus, we conduct an evaluation of argument quality across 15 dimensions, as proposed by~\citet{wachsmuth-etal-2017-argumentation}.

\textbf{Human Annotation}: We begin by selecting a subset of 25 arguments and assign raters to annotate all 15 quality dimensions. To alleviate the annotation workload, raters provide boolean judgments (high or low, rather than high/medium/low) for the first 14 dimensions, while the overall argument quality is rated using a 5-point Likert scale.

\textbf{Human-Model Agreement}: Our findings indicate that the Fine-Grained RAG Listwise LLM-Judge achieves the highest agreement with human annotations. i.e. the model's predictions of argument quality agree with human annotators when the model additionally predicts related measures like context relevance and groundedness. The results are shown in~\cref{tab:individual_llm_results_conquer}

\begin{figure*}[h]
\centering
\begin{tabular}{|>{\raggedright\arraybackslash}p{0.9\textwidth}|}
    \hline
    \\
    \texttt{You are provided with a set of 'documents' and an 'argument' that draws content from these documents. Your task is to:}
    \begin{enumerate}
        \item \texttt{Identify \{num\} sentences in the argument that directly quote or reference information from the documents, such as numbers, quotes, or the names of notable individuals or organizations. Typically, these sentences will have citations at the end, like [1], [2], etc., and will align closely with the content in the documents.}
        \item \texttt{For each identified sentence, create a modified version that contradicts the original sentence. The modified sentence should also contradict the information in the corresponding document.}
    \end{enumerate}
    \texttt{You can use various approaches to introduce contradictions. For example:}
    \begin{itemize}
        \item \texttt{The sentence "Albert Einstein developed the theory of relativity, which transformed our understanding of space and time" could be changed to "Albert Einstein developed the theory of relativity and discovered the structure of DNA, revolutionizing biology."}
        \item \texttt{Similarly, "Mount Everest is the tallest mountain in the world, standing at 8,848 meters above sea level" could be modified to "Mount Everest is the tallest mountain in the world, standing at 9,500 meters above sea level, and it is home to an ancient civilization."}
    \end{itemize}
    \texttt{Once you've made the modifications, return a new version of the argument where the identified sentences are replaced with their modified versions. All other sentences should remain unchanged.}
    \\
    \texttt{You should generate your output in a JSON format and nothing else. The output should be in JSON format, structured as follows:}
    \begin{verbatim}
    {
      "modifications": ["...", ..., ],
      "modified_argument": ".."
    }
    \end{verbatim}
    \texttt{Here are the documents and the argument:}
    \\
    \texttt{Documents: \{documents\}}
    \\
    \texttt{Original Argument: \{argument\}}
    \\
    \texttt{Ensure that your output is in a JSON format and do not generate anything else.}
    \\
    \hline
\end{tabular}
\caption{Prompt for generating modified arguments with hallucinations.}
\label{fig:modified_argument_prompt}
\end{figure*}

\section{Context Relevance Evaluation Formats}\label{sec:contextrelevanceevaluationformats}
To assess the relevance and quality of documents in relation to controversial topics, we employed a variety of prompts designed to evaluate different aspects of context relevance. These prompts guided the LLM Judges in generating ratings or structured outputs that reflect the relevance of evidence documents. Each prompt variation expects specific outputs such as numerical scores, JSON objects, or Boolean evaluations. The following subsections describe each prompt variation.

\subsection{Direct Format}
In this format, the LLM-Judge uses the prompt from a popular RAG evaluation library TruLens~\footnote{\url{https://github.com/truera/trulens/}}. The LLM is tasked to output a numerical score from 0 to 10, where 0 indicates the least relevance and 10 represents the highest relevance. The prompt provides detailed guidelines to ensure consistency in scoring, such as considering the degree of coverage of the topic by the documents, as well as the relevance to different aspects of the topic. Long and short documents are to be treated equally in terms of scoring potential.

\subsection{Fine-grained Format}
The Fine-grained Format requires the LLM-Judge to assess the relevance of documents based on specific criteria, including direct relevance, breadth of coverage, quality of evidence, applicability to argumentation, consistency, and the level of unrelated content. Each criterion is evaluated with a True or False score, accompanied by a brief explanation. The goal is to provide a detailed evaluation that identifies strengths and weaknesses in the documents. (\cref{prompt_fine_grained_context_rel})

\subsection{G-Eval Prompt}
The G-Eval Prompt focuses on evaluating documents against a set of criteria for context relevance, including direct relevance, breadth of coverage, evidence credibility, and the consistency of relevance throughout the documents. Similar to the G-Eval prompt~\cite{liu-etal-2023-g} used in other tasks, we first generate evaluation steps seperately and the LLM-Judge is expected to follow these steps to assess whether the documents collectively cover the core aspects of the controversial topic, rate for credibility, check for irrelevant content, and assign a relevance score between 1 and 5. (\cref{prompt_g_eval})

\subsection{Query-Rubric Based Format}
In the Query-Rubric Based Format, the LLM-Judge generates a list of 20 `nuggets' which are key pieces of information relevant to the given query and its stance. The relevance of these nuggets is assessed based on how well they are covered by the provided documents, with scores ranging from 1 (poorly covered) to 5 (thoroughly covered). Unlike the previous two formats, this format is interpretable and is useful in determining the degree to which the documents support or refute the query's stance by providing evidence for specific points or nuggets. Rubric based metrics as rubric based evaluation has shown success in evaluating document relevance~\cite{farzi2024pencils} and dialog~\cite{hashemi-etal-2024-llm}. (\cref{prompt_query_rubric})

The following two formats are slightly different than traditional ones as well as the ones discussed earlier. In the below two formats, we attempt to generate the context relevance, argument groundedness, and the argument's preference against a human argument in a single prompt.
\subsection{RAG-Rubric Format}
The RAG-Rubric Format extends the Query-Rubric Based approach by not only evaluating the coverage of nuggets by the documents but also comparing the coverage by two different arguments. Evaluators assess context relevance, answer relevance, answer groundedness, and argument preference. Each nugget is scored from 1 to 5 for its relevance to the context, as well as its coverage by Argument 1 and Argument 2. For assessing context relevance, we take the average score assigned to Argument 1 for each nugget (when the generated argument is placed before) and vice versa. (\cref{prompt_rag_rubric})

\subsection{RAG-Direct Format}
The Rubric-Direct Format involves evaluating two arguments (one human-written and one model-generated) on context relevance, answer relevance, answer groundedness, and argument preference. Each criterion is evaluated for both arguments, with scores assigned based on how well the arguments align with the evidence documents, adhere to the topic, and maintain consistency. The evaluation also includes determining which argument is more effective overall based on factors such as coverage of evidence, reasoning depth, coherence, and stance consistency. (\cref{prompt_rag_direct})

\begin{table*}[b]
\centering
\resizebox{0.9\columnwidth}{!}{
\begin{tabular}{c|cc|cc}
\hline
\textbf{\% Irrelevant Content} & \multicolumn{2}{c}{\textbf{Direct}} & \multicolumn{2}{c}{\textbf{RAG-Direct}} \\
 & \textbf{CON} & \textbf{PRO} & \textbf{CON} & \textbf{PRO} \\
 \hline
0 & .817 & .869 & .810 & .795 \\
10 & .703 & .797 & .779 & .755 \\
20 & .676 & .841 & .769 & .741 \\
50 & .672 & .762 & .767 & .704 \\
70 & .690 & .714 & .702 & .693 \\
100 & .210 & .231 & .529 & .453 \\
\hline
\textbf{$\rho$} & -0.82 & -0.86 & -0.903 & -0.90 \\
\begin{tabular}[c]{@{}c@{}}Monotonic\\ Decrease\end{tabular} & \xxxb & \xxxb & \chk & \chk \\
\bottomrule
\hline
\end{tabular}
}
\caption{Fine-grained metrics both monotonically decrease while single score metrics do not consistently decrease across pro as well as con types of evidence with~\texttt{gpt-4o-mini} as an evaluator.}
\label{tab:ranking_models_pro_con}
\end{table*}

\subsection{Listwise+RAG Fine-Grained Format}
The Listwise+RAG Fine-Grained Format is similar to the RAG-Direct format, with two main differences. First, the model is forced to generate a score at the granularity of each document, rather than only a collection of documents. Second, the model is also forced to generate the 15 fine-grained argumentation quality metrics. (\cref{prompt_rag_docwise})

\begin{table*}[]
\resizebox{\textwidth}{!}{
 \begin{tabular}{c|ccccccc}
    \hline
        \textbf{\% Irrelevant Context} & \textbf{Direct} & \textbf{Static-Rubric} & \textbf{G-Eval-Arg} & \textbf{Query-Rubric} & \textbf{RAG-Rubric} & \textbf{RAG-Direct} & \textbf{Listwise+RAG FG}\\ \hline
        \textbf{0} & .914 & .960 & .869 & .925 & .880 & .722 & .708\\ 
        \textbf{10} & .897 & .914 & .821 & .923 & .873 & .616 & .670\\ 
        \textbf{20} & .834 & .833 & .773 & .928 & .826 & .545 & .480\\ 
        \textbf{50} & .893 & .661 & .698 & .781 & .791 & .372 & .458\\ 
        \textbf{70} & .772 & .506 & .621 & .686 & .634 & .333 & .329\\ 
        \textbf{100} & .214 & .017 & .295 & .094 & .477 & .157 & .169\\ \hline
        $\rho \downarrow$ & -0.82 & -0.97 & -0.95 & -0.90 & -0.97 & -0.99 & -0.97\\ 
        Monotonic Decrease & \xxxb & \chk & \chk & \xxxb & \chk & \chk & \chk \\ \hline
    \end{tabular}}
\caption{\texttt{GPT-4o} Judges vs Irrelevant Context In an Oracle Setting: The context relevance predictions as we increase the amount of irrelevant context--the fine-grained metrics decrease monotonically, and are strongly negatively correlated as compared to the direct metric which emits a single score ($\rho$ = Pearson Correlation Coefficient).} 
\label{tab:context_rel}
\end{table*}

\begin{table*}[]
\resizebox{\textwidth}{!}{
\begin{tabular}{lrrrrrrr}
\hline
\multicolumn{1}{|l|}{\textbf{Metric}} & \multicolumn{1}{l|}{Direct} & \multicolumn{1}{l|}{Static-Rubric} & \multicolumn{1}{l|}{G-Eval-Arg} & \multicolumn{1}{l|}{Query-Rubric} & \multicolumn{1}{l|}{RAG-Rubric} & \multicolumn{1}{l|}{RAG-Direct} & \multicolumn{1}{l|}{Listwise+RAG FG} \\ \hline
(Gold) GPT-4o & \xxxb & \chk & \chk & \xxxb & \chk & \chk & \chk\\
(Gold) GPT-4o-mini & \xxxb & \chk & \chk & \chk & \xxxb & \chk &  -\\
(Gold) Llama-3.1-70B & \xxxb & \chk & \xxxb &  \chk &   ~\xxxb$^\gamma$ & ~\xxxb$^\gamma$ & ~\xxxb$^\gamma$\\
 (Gold) Gemini-1.5-Flash& \chk & \xxxb & \chk & \xxxb & \xxxb & \xxxb &\chk\\
\hline
(Retrieved) GPT-4o & \xxxb & \chk & \xxxb & \xxxb & \chk & \chk & \xxxb \\ 
(Retrieved) GPT-4o-mini & \xxxb & \xxxb & \chk & \chk & \xxxb & \chk & -\\
\hline
\end{tabular}
}
\caption{Presence of~\textit{strict monotonic decrease} of context relevance with increasing irrelevant context over (up) gold documents (bottom) documents retrieved through a BM25+gpt-4o-mini pipeline. $\gamma$ -- the model failed to generate a valid JSON.}
\label{tab:modelwise_context_rel}
\end{table*}
\begin{figure*}[h]
    \centering
    \includegraphics[width=1.1\textwidth]{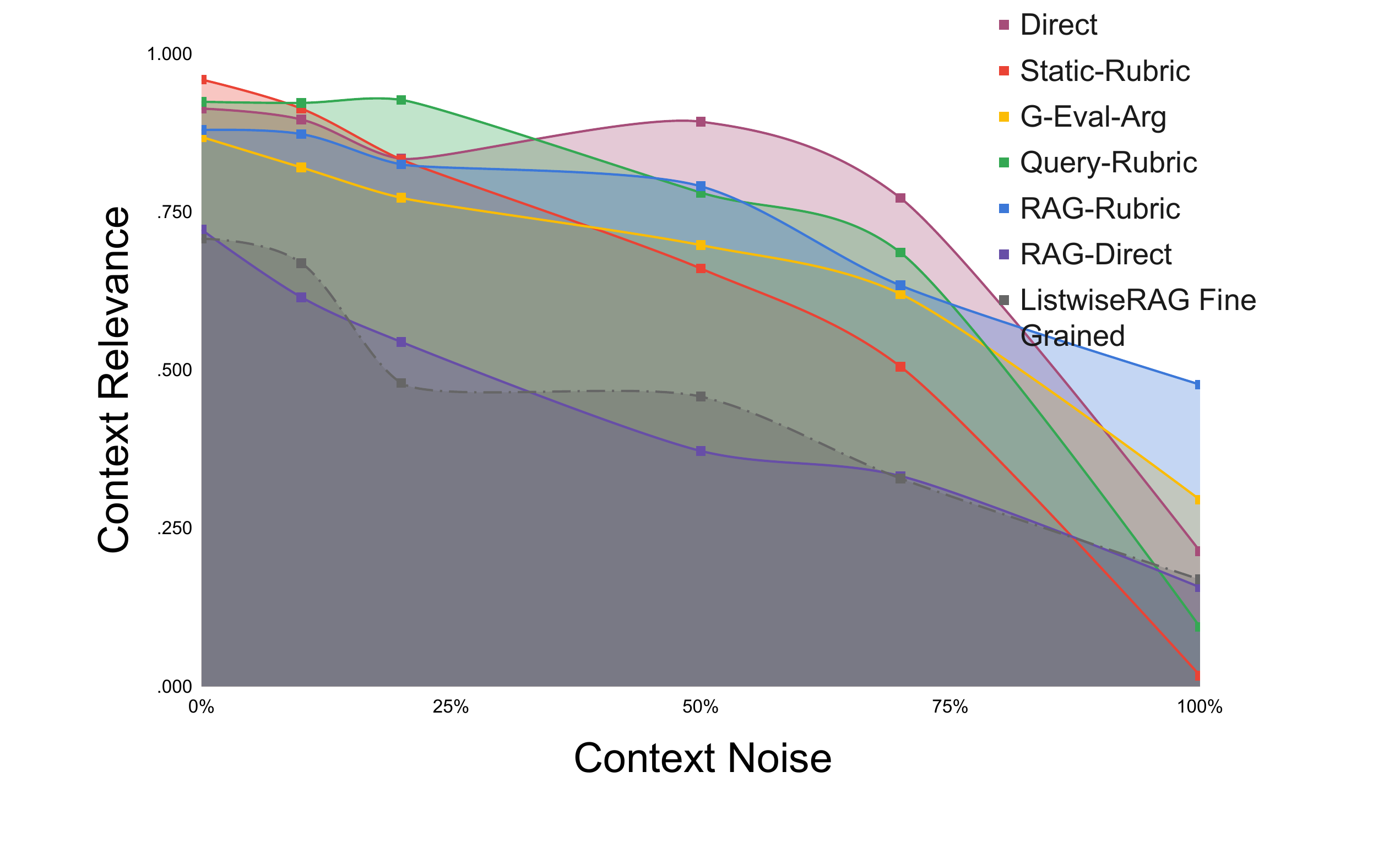}
    \caption{Sensitivity of LLM Judges predicting context relevance at different levels of irrelevance}
    \label{fig:context_relevannce_sensitivity_front}
\end{figure*}

\begin{table*}[h!]
\centering
\resizebox{\textwidth}{!}{
\begin{tabular}{c|rrrrrrr}
\toprule
\hline
\begin{tabular}[c]{@{}c@{}}Number of Argument\\ Sentences Hallucinated\end{tabular} & \multicolumn{1}{l}{\textbf{Direct}} & \multicolumn{1}{l}{\textbf{Static-Rubric}} & \multicolumn{1}{l}{\textbf{G-Eval-Arg}} & \multicolumn{1}{l}{\textbf{Query-Rubric}} & \multicolumn{1}{l}{\textbf{RAG-Rubric}} & \multicolumn{1}{l}{\textbf{RAG-Direct}} & \multicolumn{1}{l}{\textbf{Listwise+RAG FG}} \\
\hline
0 & .862 & .778 & .825 & .914 & .899 & .681 & .979\\
5 & .552 & .224 & .491 & .682 & .627 & .462 & .938\\
20 & .193 & .0 & .266 & .210 & .212 & .0 & .729\\
\hline
\bottomrule
\end{tabular}
}
\caption{Argument Groundedness at varying levels of hallucinated content.}
\label{tab:answer_groundedness}
\end{table*}

\end{document}